\documentclass[sigconf]{acmart}
\usepackage{multirow}
\usepackage{microtype}
\usepackage{graphicx}
\usepackage{subfigure}
\usepackage{url}
\usepackage{booktabs, paralist}
\usepackage[utf8]{inputenc}
\usepackage{listings}
\usepackage{caption}
\usepackage{xcolor}
\usepackage{dsfont}
\usepackage{algorithm}
\usepackage{algpseudocode}

\usepackage{amsmath,amsfonts,bm}

\def\eqref#1{equation~\ref{#1}}

\def\1{\bm{1}}

\def\ve{{\bm{e}}}

\def\vh{{\bm{h}}}

\def\vm{{\bm{m}}}

\def\vp{{\bm{p}}}
\def\vq{{\bm{q}}}
\def\vr{{\bm{r}}}
\def\vs{{\bm{s}}}

\def\mA{{\bm{A}}}

\def\mD{{\bm{D}}}

\def\mI{{\bm{I}}}

\def\mM{{\bm{M}}}

\def\mQ{{\bm{Q}}}

\def\mY{{\bm{Y}}}

\DeclareMathAlphabet{\mathsfit}{\encodingdefault}{\sfdefault}{m}{sl}
\SetMathAlphabet{\mathsfit}{bold}{\encodingdefault}{\sfdefault}{bx}{n}

\def\gG{{\mathcal{G}}}

\def\sG{{\mathbb{G}}}

\newcommand{\R}{\mathbb{R}}

\DeclareMathOperator*{\argmax}{arg\,max}

\definecolor{codegreen}{rgb}{0,0.6,0}
\definecolor{codegray}{rgb}{0.5,0.5,0.5}
\definecolor{codepurple}{rgb}{0.58,0,0.82}
\definecolor{backcolour}{rgb}{0.95,0.95,0.92}

\lstdefinestyle{style_full}{
    commentstyle=\color{codegreen},
    keywordstyle=\color{magenta},
    numberstyle=\tiny\color{codegray},
    stringstyle=\color{codepurple},
    basicstyle=\ttfamily\footnotesize,
    breakatwhitespace=false,         
    breaklines=true,                 
    captionpos=b,                    
    keepspaces=true,                 
    numbers=left,                    
    numbersep=8pt,                  
    showspaces=false,                
    showstringspaces=false,
    showtabs=false,                  
    tabsize=4,
    xleftmargin=3em
}

\lstdefinestyle{style_snippet}{
    commentstyle=\color{codegreen},
    keywordstyle=\color{magenta},
    numberstyle=\tiny\color{codegray},
    stringstyle=\color{codepurple},
    basicstyle=\ttfamily\scriptsize,
    breakatwhitespace=false,         
    breaklines=true,                 
    captionpos=b,                    
    keepspaces=true,                 
    numbers=left,                    
    numbersep=8pt,                  
    showspaces=false,                
    showstringspaces=false,
    showtabs=false,                  
    tabsize=4,
    xleftmargin=3em
}

\newenvironment{sequation}{\begin{equation}\small}{\end{equation}}
\newcommand{\method}{MICRO-Graph}
\makeatletter
\DeclareRobustCommand*\cal{\@fontswitch\relax\mathcal}
\makeatother
\AtBeginDocument{%
  \providecommand\BibTeX{{%
    \normalfont B\kern-0.5em{\scshape i\kern-0.25em b}\kern-0.8em\TeX}}}

\setcopyright{acmcopyright}
\copyrightyear{2021}
\acmYear{2021}
\acmDOI{10.1145/1122445.1122456}

\acmConference[Ljubljana '21]{}{April 19-23, 2021}{Ljubljana, Slovenia}

\setcopyright{none}
\begin{document}

\title{Motif-Driven Contrastive Learning of Graph Representations}

\author{Shichang Zhang}
\authornote{Both authors contributed equally to this research.}
\affiliation{%
  \institution{University of California, Los Angeles}
  \state{CA}
  \country{USA}
}
\email{shichang@cs.ucla.edu}

\author{Ziniu Hu}
\authornotemark[1]
\affiliation{%
  \institution{University of California, Los Angeles}
  \state{CA}
  \country{USA}
}
\email{bull@cs.ucla.edu}

\author{Arjun Subramonian}
\affiliation{%
  \institution{University of California, Los Angeles}
  \state{CA}
  \country{USA}
}
\email{arjunsub@ucla.edu}

\author{Yizhou Sun}
\affiliation{%
  \institution{University of California, Los Angeles}
  \state{CA}
  \country{USA}
}
\email{yzsun@cs.ucla.edu}


\begin{abstract}

Pre-training Graph Neural Networks (GNN) via self-supervised contrastive learning has recently drawn lots of attention. 
However, most existing works focus on node-level contrastive learning, which cannot capture global graph structure. 
The key challenge to conducting subgraph-level contrastive learning is to sample informative subgraphs that are semantically meaningful. 
To solve it, we propose to learn graph motifs, which are frequently-occurring subgraph patterns (e.g. functional groups of molecules), for better subgraph sampling. Our framework \underline{M}ot\underline{I}f-driven \underline{C}ontrastive lea\underline{R}ning \underline{O}f \underline{G}raph representations (\textit{\method}) can: 1) use GNNs to extract motifs from large graph datasets; 2) leverage learned motifs to sample informative subgraphs for contrastive learning of GNN.
We formulate motif learning as a differentiable clustering problem, and adopt EM-clustering to group similar and significant subgraphs into several motifs. Guided by these learned motifs, a sampler is trained to generate more informative subgraphs, and these subgraphs are used to train GNNs through graph-to-subgraph contrastive learning. 
By pre-training on the ogbg-molhiv dataset with  \textit{\method}, the pre-trained GNN achieves 2.04$\%$ ROC-AUC average performance enhancement on various downstream benchmark datasets, which is significantly higher than other state-of-the-art self-supervised learning baselines. 
\end{abstract}





\maketitle

\section{Introduction}\label{sec:introduction}
Graph-structured data, such as molecules and social networks, are ubiquitous in many scientific research areas and real-world applications. Recently, Graph Neural Networks (GNNs) have shown great expressive power for learning graph representations without explicit feature engineering \cite{kipf2016semi, hamilton2017inductive, velivckovic2017graph, xu2018powerful}. To empower GNNs to capture graph structural and semantic properties without human annotations, a line of works has been proposed to pre-train GNNs in a self-supervised manner \cite{velikovi2018deep, gpt_gnn, qiu2020gcc, bai2019unsupervised, navarin2018pretraining, Wang2020Inductive, sun2019infograph, hu2020pretraining, you2020graph}. The pre-trained GNNs could generalize to downstream tasks on the graphs within the same domain (e.g., all molecules in the chemical domain) and enhance performance with a few fine-tuning steps.

\begin{figure}[t]
\begin{center}
\includegraphics[width=0.95\columnwidth]{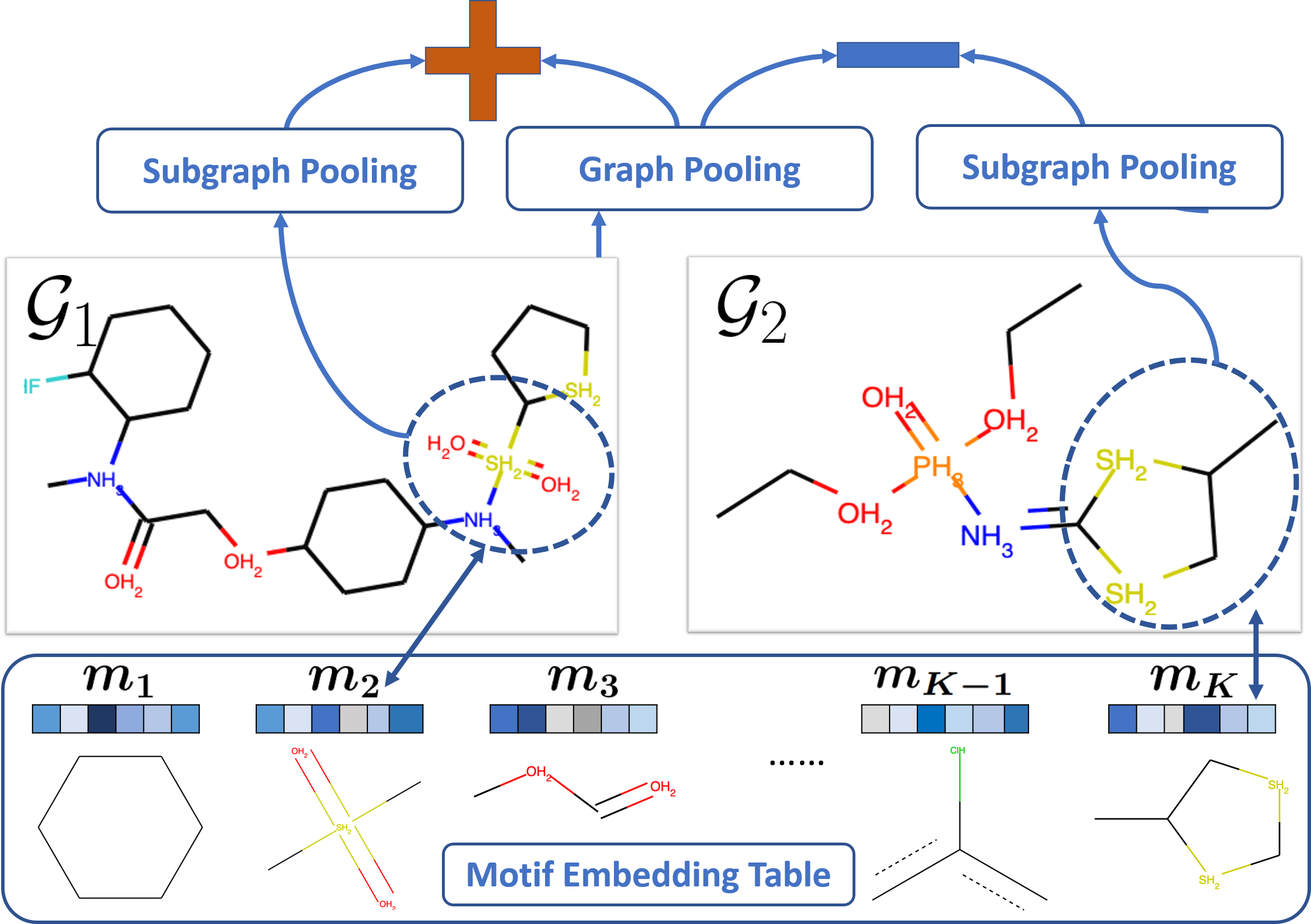} 
\end{center}
\caption{Given a graph dataset, we learn a motif embedding table storing prototypical embeddings of motifs. For a pair of input graphs $\gG_1$ and $\gG_2$, we leverage learned motifs to generate motif-like subgraphs and conduct graph-to-subgraph contrastive learning.}
\label{figure:key_idea}
\vskip -0.2in
\end{figure}

Many GNN pre-training works could be categorized into the contrastive learning framework, which forces views from the same data instance (e.g., different crops from an image or different nodes from a graph) to become closer and pushes views from different instances apart. One key component in contrastive learning is to generate informative and diverse views from each data instance. For example, in computer vision, researchers use various augmentation methods, including cropping, color distortion, and Gaussian blurs, to generate image views~\cite{chen2020simple}. However, due to graph structure's discrete nature, constructing informative views of a graph is a challenging task. Most existing works~\cite{velikovi2018deep, sun2019infograph, hu2020pretraining} utilize nodes as views for contrastive learning, which lacks the ability to guide GNNs to capture the global information of graphs during pre-training, and thus limits the performance enhancement during fine-tuning on downstream tasks. On the other hand, even though subgraphs are higher-level views superior to nodes for capturing global information, sampling informative subgraphs for contrastive learning is a non-trivial task. Existing sampling techniques such as random walk and k-hop neighbors are non-ideal because they only consider local structures and overlook node features, resulting in node chains or rings, which are not semantically meaningful a subgraph.

To study the characteristics of meaningful subgraphs, researchers in the graph mining community have proposed to uncover global properties of graphs through graph motifs, which are defined as significant subgraph patterns that frequently occur \cite{milo2002network}. For example, hydroxide (–$OH$), a functional group, usually implies higher water solubility of small molecules, and $Zif268$, a protein structure, can mediate protein-protein interactions in sequence-specific DNA-binding proteins. \cite{pabo2001design}. 
Due to motifs' great expressiveness, we propose to learn motifs from a given graph dataset and leverage learned motifs to sample informative subgraphs for GNN contrastive learning.

However, existing motif mining techniques could not be utilized directly to serve our purpose because they rely on discrete counting of subgraph structures \cite{milo2002network, kashtan2004efficient, chen2006nemofinder, wernicke2006efficient}. This limitation makes it hard to generalize large-scale graph datasets with continuous and high-dimensional node features, as is often the case in real-world applications.

In light of the significance and challenges of motif learning, we propose \textit{\method}: a framework for \underline{M}ot\underline{I}f-driven \underline{C}ontrastive lea\underline{R}ning \underline{O}f \underline{G}raph representations to: 1) use GNNs to automatically extract graph motifs as prototypical subgraph embeddings from large graph datasets; 2) leverage the learned graph motifs to generate informative motif-like subgraphs to benefit contrastive learning of GNNs. The motif learning module and the contrastive learning module are mutually reinforced to train a more expressive GNN encoder that can extract meaningful motifs. An illustration of this key idea is shown in Figure \ref{figure:key_idea}.

For motif learning, we tackle the challenging discrete graph motif mining problem by representing motifs as continuous embeddings, so the framework becomes differentiable. Specifically, we encode sampled motif-like subgraphs via a GNN encoder and get subgraph embeddings. Then, we treat graph motifs as a latent variable and learn them by maximizing the graph likelihood through an EM algorithm.

For contrastive learning, we tackle the challenge of informative subgraph generation by leveraging learned motifs as guidance to generate motif-like subgraphs. Specifically, we partition all nodes in a graph and induce subgraphs with a high probability of belonging to a specific motif. As motifs represent the critical graph properties by nature, the motif-like subgraphs are more informative than graph nodes and randomly sampled subgraphs, thus enhancing the contrastive learning to capture global graph characteristics. 

The pre-trained GNN using \textit{\method} on the ogbg-molhiv molecule dataset can successfully learn some meaningful functional groups as motifs, including Benzene rings, nitro, acetate, etc., which help interpret the model decisions. Meanwhile, fine-tuning this GNN on seven chemical property prediction benchmarks yields 2.04\% ROC-AUC average improvement over non-pretrained GNNs and outperforms other state-of-the-art GNN pre-training baselines. Extensive ablation studies show the important role of motif learning in our framework.

We summarize the contributions of this paper as follows:
\begin{compactitem}
\item Utilize graph motifs to generate more informative subgraphs to improve contrastive GNN pre-training.
\item Turn the discrete and non-scalable motif learning problem into differentiable so that we can extract significant motifs from large graph datasets with rich features.
\item Achieve the best results on various chemical property prediction tasks than existing GNN pre-training techniques, and the learned motifs can facilitate researchers to interpret model decisions and scientific discovery.
\end{compactitem}

\section{Related Work}\label{sec:related}

\textbf{Contrastive learning} is a widely-used self-supervised learning algorithm, which recently achieves great results for visual representation learning \cite{chen2020simple, he2019moco, DBLP:journals/corr/abs-1905-09272}. 
One key component in contrastive learning is to generate informative and diverse views from each data instance. In computer vision, early researchers leveraged a pixel as a local view to conduct local-to-local \cite{DBLP:journals/corr/abs-1807-03748} or local-to-global contrastive learning \cite{DBLP:conf/iclr/HjelmFLGBTB19, DBLP:conf/nips/BachmanHB19}, while recently, researchers have found that randomly-cropped image snippets~\cite{chen2020simple, he2019moco} help a model to better capture the relationships between image elements. This motivates us to conduct contrastive learning of GNNs at the subgraph level.

\textbf{Contrastive learning for GNNs} has drawn much attention recently. Most existing works utilize a node as a view to conduct node-to-context~\cite{hu2020pretraining, gpt_gnn} or node-to-graph contrastive learning~\cite{velikovi2018deep, sun2019infograph}. Similar to the limitation of using a pixel as a local view, GNNs trained via node-level contrastive learning can only utilize local graph structure to determine whether a node belongs to a graph, but they are less effective at capturing whole-graph characteristics. To tackle this limitation, several works try to conduct contrastive learning on subgraphs, which are sampled via heuristic strategies like random walk~\cite{qiu2020gcc, you2020graph}. However, as these heuristic sampling strategies are random and only consider graph structures but not features, the sampled subgraphs are prone to be non-meaningful. For example, utilizing random walk on molecules is likely to generate a chain graph, which is not very helpful for contrastive learning. To tackle this limitation, we propose to leverage graph motifs, i.e., significant and frequently-occurring subgraph patterns, to guide informative subgraph generation for contrastive learning. 

A recent GNN pre-training work Grover~\cite{rong2020selfsupervised} also proposes to use motifs as self-supervision. The main difference is that they utilize traditional software to extract discrete motifs and treat them as node classification labels. Our work instead learns motifs through joint pre-training, and we leverage the motifs, which could encode richer graph semantics, to benefit contrastive learning.

\paragraph{Graph motifs} are frequently-occurring subgraph patterns, which are the building blocks of complex graphs. 
Mining motifs can benefit many tasks from exploratory analysis to transfer learning \cite{henderson2012rolx}. For many years, various motif mining algorithms have been proposed. There are generally two categories, either (1) exact counting as in \cite{milo2002network, kashtan2004efficient, schreiber2005frequency, chen2006nemofinder}, or (2) sampling and statistical estimation as in \cite{wernicke2006efficient}. However, both approaches cannot scale to large graph datasets with high-dimensional and continuous features, which is common in real-world applications. In this paper, we propose to turn the discrete motif-learning problem into differentiable, so that the framework can learn motifs from large-scale graph datasets with rich node features.

\section{Method} \label{sec:methodology}
\subsection{Motif-Driven Contrastive Learning Framework} \label{subsec:framework}
The problem we want to solve is to pre-train a GNN encoder \textbf{ENC}$_{\theta}(\cdot)$ to capture significant characteristics of input graph data in a self-supervised manner (i.e., without human annotations), where $\theta$ denotes the parameters of the encoder. The pre-trained GNN can encode graphs in the same domain to $d$-dimensional embeddings capturing their fundamental semantic properties. They can be generalized to various downstream graph tasks, e.g., chemical graph property prediction, even with few labeled data for fine-tuning.

To guide the GNN to capture global graph characteristics, we use subgraph-level contrastive learning as the self-supervised objective. The major challenge is to sample semantically-informative subgraphs. We propose to learn graph motifs via a differentiable update
and leverage the learned motifs to sample informative subgraphs to tackle it. Formally, we represent motifs in a continuous embedding space as a $K$-slot Motif Embedding Table $\mM = \{\vm_{1}, \cdots, \vm_{K}\}$.  Each slot stores a continuous $d$-dimensional motif embedding, which is a prototypical representation of subgraph embeddings. Guided by the learned motif embeddings $\mM$, we group nodes to form subgraphs with a high probability of belonging to a specific motif. These motif-like subgraphs are informative and can benefit contrastive learning.

\begin{figure*}[h!]
\begin{center}
\includegraphics[width=\textwidth]{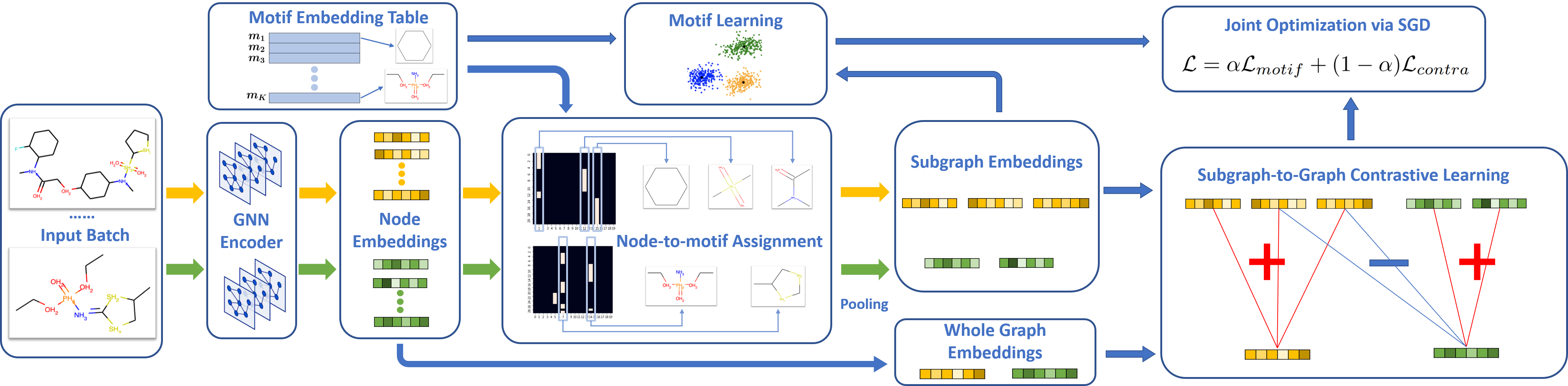} 
\end{center}
\caption{Overall framework of \textit{\method}. A GNN trained in a self-supervised manner to automatically extract motifs. The learned motifs are leveraged to generate informative subgraphs for graph-to-subgraph contrastive learning. }
\label{figure:framework}
\end{figure*}



To learn both the GNN encoder \textbf{ENC}$_{\theta}$ and the Motif Embedding Table $\mM$ jointly, we propose a differentiable learning framework \textit{\method}, as is illustrated in Figure \ref{figure:framework}. We first pass a batch of graphs into the GNN encoder \textbf{ENC}$_{\theta}$ to get their contextualized node embeddings. Then, we group nodes to sample motif-like subgraphs, which are further pooled to get subgraph embeddings.
Afterward, we feed the subgraph embeddings into the two learning modules. The \textit{Motif-Learner} updates the motif embeddings $\mM$ by maximizing the likelihood of these subgraphs, and the \textit{Contrastive Learning} module updates GNN parameters $\theta$. 
We introduce each module in detail in the following sections.






\subsection{Differentiable Motif Learning}\label{subsec:motif_learner}
We first introduce how to automatically learn motif embeddings $\mM$, as well as leverage them to generate informative motif-like subgraphs in a fully-differentiable manner. This process contains two coupled optimization problems: (1) given motif embeddings $\mM$, how do we partition a graph into a set of subgraphs that are most similar to the prototypical motifs; (2) given the set of subgraphs, how can we update $\mM$ and get better prototypical motifs?

We derive a probabilistic model by drawing an analogy to the PLSA topic model\cite{hofmann2013probabilistic}, where motifs correspond to topics. Given a graph $\gG$ with $N$ nodes,  we generate $N$ node embeddings via the GNN encoder, i.e. $\{\vh_{1}, \cdots, \vh_{N}\} = $ \textbf{ENC}$_{\theta}(\gG)$, and we denote a partition of these nodes as $\bm{Par}$. Each element in $\bm{Par}$ is a set of nodes, which can form an induced subgraph in $\gG$. When aggregate the node embeddings in each induced subgraph together, we get the subgraph embedding $\vs$. In shorthand, we denote the procedure of generating subgraph embeddings in $\gG$ through partition and aggregation as $\{\vs_1, \cdots, \vs_{J}\} = \gG[\bm{Par}]$. To model the likelihood of generating these subgraphs from motifs, we define a K-way categorical random variable $z_j \in \{1, \cdots, K\}$ for each subgraph, with $P(\vs_j \vert z_j = k)$ representing the probability of generating $\vs_j$ from the $k$-th motif. Then the likelihood of a subgraph is the following:
\begin{align}\label{equation:sub_likelihood}
P(\vs_j \vert \mM, \theta) = \sum_{k=1}^{K} P(\vs_j \vert z_j = k, \mM, \theta) P(z_j = k)
\end{align}


The conditional likelihood of the whole graph $\gG$ given a partition $\bm{Par}$ is the following: 
\begin{align}\label{equation:conditioanl_likelihood}
&P(\gG \vert \bm{Par}, \mM, \theta) \nonumber \\
&= \prod_{\vs_j \in \gG[\bm{Par}]} \sum_{k=1}^{K} P(\vs_j \vert z_j = k, \mM, \theta) P(z_j = k)
\end{align}

Our first learning objective is to update the motif embedding $\mM$ to maximize the conditional likelihood in Eq (\ref{equation:conditioanl_likelihood}). This is a standard clustering problem involving hidden variables $z$ and can be optimized via an EM algorithm.

The next problem is to find the optimal partition $\bm{Par}^*$ that maximizes the  likelihood $P(\gG \vert \bm{Par}, \mM, \theta)$. This task is very challenging as each partition $\bm{Par}$ is a combinatorial set over all nodes and the search space grows exponentially in the graph size. It would be computationally-intensive to enumerate $P(\vs_j \vert z_j = k)$ for all $\vs_j$ under all possible partitions. To alleviate this issue, we propose to approximate $P(\vs_j \vert z_j = k)$ by breaking it down to the node level. Specifically, we define another $K$-way categorical random variable $c_l$ for each node, where $P(\vh_l \vert c_l = k)$ representing the probability of node $\vh_l$ being generated from the $k$-th motif, which is analogous to the word distribution for each topic. 

Assuming conditional independence between nodes given the subgraph they belong to and the motif they come from, we derive a likelihood of Eq (\ref{equation:conditioanl_likelihood}) using node embeddings as the following:
\begin{align}\label{equation:likelihood_partition}
&\hat{P}(\gG \vert \bm{Par}, \mM, \theta) \nonumber = \\
&\prod_{\vs_j \in \gG[\bm{Par}]} \sum_{k=1}^{K} \prod_{\substack{{l=1} \\ \vh_l \in \vs_j }}^{N} P(\vh_l \vert c_l = k, \mM, \theta) P(z_j = k)
\end{align}
For this objective, we can also optimize it via an EM algorithm. In the E-step we calculate the posterior probability $\vq_{l, k} = P(c_l = k \vert \vh_l)$. In the M-step, we optimize the objective to find the optimal partition and parameter $\theta$. We discuss more about how EM steps work exactly below and defer the detailed mathematical derivations to Appendix \ref{appendix:theory}.


To summarize, learning motifs involves solving two coupled optimization problem sequentially as shown below, where we based on the current $\mM$ to optimize over partitions and GNN parameters, and then optimize $\mM$ given the the optimal partition and GNN parameters.
\begin{align}
    \bm{Par}^{*},\ \theta^{*} &= \arg\max_{\bm{Par}, \theta} \hat{P}(\gG | \bm{Par}, \mM, \theta)\label{eq:partition} \\
    \mM^{*}  &= \arg\max_{\mM} P(\gG | \bm{Par}^{*}, \mM, \theta^{*}) \label{eq:motif}
\end{align} 
Notice that the first objective doesn't optimize the motif embeddings $\mM$, as the node-to-motif probability is just an approximation, and we would like to use subgraph embeddings for learning $\mM$ in the second objective, which only optimizes $\mM$.


We show how to parameterize these two components and the learning details in the following sections.

\paragraph{Modeling and Learning for The Graph Partition.}
We first introduce how we model and optimize the graph partition problem in Eq (\ref{eq:partition}) via an EM algorithm. We model $\vq_{l, k}$ using the similarity between node embeddings and motif embeddings. To be able to measure the similarity, we map $\vh_{l}$ with a projection parameter matrix $W_h$ to the motif embedding space. We then compute the pairwise cosine similarities between nodes and motifs, followed by a softmax normalization to turn the values into probabilities. 
\begin{equation} \label{equation:node_motif}
\vq_{l, k} = \frac{\exp\big(\phi(W_h \vh_{l})^T\phi(\vm_{k}) / \tau\big)}{\sum_{k'}\exp\big(\phi(W_h \vh_{l})^T\phi(\vm_{k'}) / \tau\big)}
\end{equation}
Here we use $\phi(x) = x \ / \ \Vert x \Vert_{2}$ to denote L-2 normalization, and $\tau$ is a temperature hyper-parameter. Given a batch of graphs  $\sG = \{\gG_{1}, \cdots, \gG_{B} \}$, we use $\mQ = [\vq^{(1)}, \cdots, \vq^{(B)}]^T$ to denote the corresponding node-to-motif probabilities for all $N_B$ nodes in a batch of $B$ graphs. 


In the E-step of a standard EM algorithm, we calculate the posterior probability $P(c_l = k \vert \vh_l)$ as the reference distribution to derive the lower bound. However, directly calculate it via Eq (\ref{equation:node_motif}) and plug it into the M-step could be problematic, as there exists a degenerate solution in which all embeddings collapse together and get assigned to the same motif. Similar issues were observed in previous works when researchers were performing representation learning and assignment estimation together \cite{asano2020self, caron2020unsupervised}. The degenerate problem comes from extreme assignments of nodes. To avoid this issue, we adopt the strategy used in \citet{asano2020self} to derive $\hat\mQ$ as a more balanced estimate of $\mQ$ by solving a regularized optimization problem as in Eq (\ref{equation:sinkhorn_obj}) and discretize $\hat\mQ$ by only keeping the maximum column \cite{asano2020self}. We denote the discretized optimal $\hat\mQ$ as $\mQ^*$.
\begin{gather} \label{equation:sinkhorn_obj}
    \max_{\bm\hat\mQ \in \cal{Q}} Tr(\bm\hat\mQ \mQ^T) + \frac{1}{\lambda} H(\bm\hat\mQ) \text{,\  where}\\
    {\cal Q} = \{\bm\hat\mQ \in \mathbb{R}_+^{N_B, K} | \bm\hat\mQ \bm{1_{K}} = \frac{\bm{1_{N_{B}}}}{N_B}, \bm\hat \mQ^T \bm{1_{N_{B}}} = \frac{\bm{1_K}}{K}\} 
\end{gather}
Here $H(\bm\hat \mQ) = - \sum_{l,k}\bm\hat\mQ_{l,k} \log \bm\hat\mQ_{l,k}$ stands for entropy, $\bm{1_{N}}$ and $\bm{1_{K}}$ are all one vectors to force the motif assignments to be balanced. This constrained optimization problem can be solved efficiently using a fast Sinkhorn-Knopp algorithm as shown in \citet{cuturi2013sinkhorn}.


Then we use the $\mQ^*$ instead of $\mQ$ as an estimation of the posterior probability. We derive the optimal partition $\bm{Par}^*$ by maximizing Eq (\ref{equation:likelihood_partition}), which can be approximated as the following (derivation in Appendix \ref{appendix:theory})
\begin{align} \label{equation:subgraph}
\bm{Par}^* = \arg\max_{\bm{Par}}\prod_{\vs_j \in \gG[\bm{Par}]} \sum_{k=1}^{K} \prod_{\substack{{l=1} \\ \vh_l \in \vs_j }}^{N} \vq_{l, k}^* 
\end{align}

As $\vq_{l, k}^*$ is either zero or one, the assignment of each node is certain. This helps us to simplify the challenging $\bm{Par}$ optimization problem. If a subgraph contains nodes with different motif assignments, the total product will be zero. Therefore, one optimal solution is simply grouping all the nodes belonging to the same motif as a subgraph. As we add the balanced-assignment regularization, this solution can nicely partition the graph into multiple subgraphs that have a high probability of belonging to a specific motif, which matches our design purpose. In practice, we also include randomness by adding small node perturbation to the derived partition. We thus get both informative and also diverse subgraphs.

Additionally, we update the GNN parameters $\theta$ to maximize the likelihood of generating the optimal partition. Given the calculated assignment $\mQ^*$ in the E-step, we also consider the negative log-likelihood loss for updating $\theta$ as the following.
\begin{align} \label{equation: node_lower_bound}
{\cal L}_{node\text{-}mot} &= - \sum_{l=1}^{N} \sum_{k=1}^{K} \vq_{l, k}^* \log \vq_{l, k}
\end{align}

One potential improvement of this partition method is to add another regularization term to include graph structural information explicitly. Since in the current setting we don't constrain the nodes in sampled subgraphs to be connected, and the graph structure information is only used implicitly by generating node embeddings using GNN. This could result in an assignment that relies too much on feature information but overlooks structural information. To capture structural information more explicitly, we include another, spectral clustering-based regularization term ${\cal L}_{reg}$ proposed by \citet{bianchi2020spectral}:

\begin{equation}  \label{eq: mincut}
    {\cal L}_{reg} \ = -\frac{Tr(\vq^T \mA \vq)}{Tr(\vq^T \mD \vq)} + \bigg\Vert\frac{\bm\tilde\vq^T\bm\tilde\vq}{\Vert\bm\tilde\vq^T\bm\tilde\vq\Vert_{F}} - \frac{\mI_{J}}{\sqrt{J}}\bigg\Vert_{F}
\end{equation}
where $||\cdot||_{F}$ denotes the Frobenius norm, $\mA$ and $\mD$ are the adjacency and degree matrix of $\gG$, and $\bm\tilde\vq$ is $\vq$ with only $J$ selected columns, i.e. only the motifs we are producing corresponding subgraphs. This regularization guides GNN to make $\bm\hat\mQ$ closer the result of spectral clustering, and penalizes disconnected nodes being assigned to the same motif, thereby balancing the usage of feature and structural information in motif-like subgraph generation. We experiment and discuss more about this extra term in the ablation study section \ref{subsec:ablation}.

\paragraph{Modeling and Learning for Motif Embeddings.}
We now introduce how given the optimal partition we model and optimize the motif embeddings in Eq (\ref{eq:motif}). 

After we get the partition $\bm{Par}^*$ with $J$ subgraphs, we aggregate the node embeddings to get induced subgraph embeddings $\{\vs_{j}\}_{j=1}^{J}$.
Similar to the node-to-motif assignment, we estimate the probability $P(z_j = k \vert \vs_j) = \vp_{j, k}$ as the normalized similarity between the motif embeddings and the projected subgraph embeddings via $W_s$
\begin{equation} \label{equation:sub_motif}
\vp_{j, k} = \frac{\exp(\phi(W_s \vs_{j})^T\phi(\vm_{k}) / \tau)}{\sum_{j'}\exp(\phi(W_s \vs_{j'})^T\phi(\vm_{k}) / \tau)}
\end{equation}
As we mentioned above, the objective in Eq (\ref{equation:sub_motif}) now becomes a standard clustering problem like the PLSA model. We thus solve it with another standard EM algorithm. For E-step of calculating the subgraph assignment $\bm \pi_{j, k}$, we directly infer that with our result $\mQ^*$ from the node level, because subgraph sampled by our partition operation only contains nodes belonging to the same motif. We thus set $\bm \pi_{j, k} = \mQ_{l, k}^*, \forall \vh_l \in \vs_j$. Based on this, we update the motif embeddings $\mM$ by maximizing the data likelihood as the following:
\begin{align}
{\cal L}_{mot\text{-}sub} &= - \sum_{j=1}^{J} \sum_{k=1}^{K} \bm \pi_{j, k} \log \vp_{j, k}
\end{align}



As all the modules that in the loss functions are differentiable, we can accumulate all the losses together as Eq (\ref{equation:motif_total_loss}) and use a gradient-based optimizer to update both the GNN parameters $\theta$ and motif embeddings $\bm M$ jointly.
\begin{align}
{\cal L}_{motif} = 
    \lambda_n {\cal L}_{node\text{-}mot} + \lambda_s {\cal L}_{mot\text{-}sub} + \lambda_r {\cal L}_{reg} \label{equation:motif_total_loss}
\end{align}

\begin{algorithm}[!h]
\lstset{style=style_snippet}
\begin{lstlisting}[language=Python]
enc = GNN(args) # Any GNN Model could be used
M = nn.Parameters(K, embed_dim) # Motif Table
def forward(data)
    h, e = enc(data) #node and graph embeddings
    # Get Motif-like Subgraphs via Partition
    Q = cos_sim(W_h*h, M.detach()).softmax()
    with torch.no_grad():
        # Don't store gradient for discrete ops
        Q_hat = sinkhorn(Q)
        s, P_hat, num_subs = pool_sub(h, Q_hat)
    # Calculate the two loss for joint learning
    l_m = motif_loss(Q, Q_hat, data.adj, s, P_hat)
    l_c = contrastive_loss(s, e, num_subs)  
    return l_m + l_c
def motif_loss(Q, Q_hat, adj, s, P_hat):
    # Calculate motif-to-subgraph score
    P = cos_sim(W_s*s.detach(), M).softmax()
    # Calculate the two loss via M-step
    loss_mot_sub  = -(P_hat * P.log()).mean()
    loss_node_mot = -(Q_hat * Q.log()).mean()
    loss_reg      = spectral_loss(Q, adj)
    return loss_mot_sub + loss_node_mot + loss_reg
def contrastive_loss(s, e, num_subs):
    # Force pairs from the same graph closer
    Y_lab = block_diag(num_subs)
    Y = cos_sim(W_e * e, s).softmax()
    return -(Y_lab * Y.log()).mean()
\end{lstlisting}
\caption{PyTorch code, Full Version in Appendix \ref{appendix:code}}
\vskip -0.1in
\label{algo:main}
\end{algorithm}

\subsection{Motif-guided Contrastive Learning} \label{subsec:contrastive}
Based on the informative sampled motif-like subgraphs, we can train the GNN encoder \textbf{ENC}$_{\theta}(\cdot)$ via contrastive learning on the subgraph level. Specifically, we denote the graph embeddings of $\sG = \{\gG_{1}, \cdots, \gG_{B}\}$ as $\{\ve_1, \cdots, \ve_B\}$. For all $J_B = \sum_{i=1}^{B} J_i$ subgraphs sampled from this batch, we compute the graph-to-subgraph similarity matrix $\mY \in \R^{B \times J_B}$, where the entry $(i, j)$ is the cosine similarity between the corresponding graph-subgraph pair, i.e. $\mY_{i, j} = \phi(W_e \ve_{i})^T \phi(\vs_{j})$, where $W_e$ is a projection parameter matrix. For each graph $\gG_i$, subgraphs sampled from it are considered as positive pairs to it, while subgraphs from other graphs are considered as negative pairs. Thus, the contrastive objective function is calculated as follows:
\begin{sequation}
    {\cal L}_{contra}  = -\frac{1}{B} \sum_{i=1}^{B} \sum_{\vs_j \in \gG_i} \log \frac{\exp(\mY_{i, j} / \tau)}{\sum_{j'}\exp(\mY_{i, j'} / \tau)} 
\end{sequation}




\subsection{MICRO-Graph Joint Training}
The whole framework of \textit{\method} can be trained jointly with a weighted sum of the two loss described above.

\begin{equation} \label{equation:joint_loss}
{\cal L} =  \alpha {\cal L}_{motif} + (1 - \alpha) {\cal L}_{contra}
\end{equation}

The overall training details are depicted in Algorithm \ref{algo:main}. The motif learning and contrastive learning modules mutually enhance each other. Initially, the motif embeddings are randomly initialized, and the motif-like subgraphs are random. As the training proceeds, a better-trained GNN can help generate more representative subgraph embeddings for motif learning, while more informative subgraphs can benefit contrastive learning.

\section{Experiments}\label{sec:evaluation}
\begin{table*}[h]
\small
\center
\caption{Transfer Fine-tune performance (ROC-AUC) of \textit{\method} compared with other self-supervised learning (SSL) baselines on molecule property prediction benchmarks. Pre-train GNNs on ogbg-molhiv dataset, fine-tune on each downstream task for ten times.}
\resizebox{\textwidth}{!}{
\begin{tabular}{@{}l|ccccccc|c@{}}
\toprule
\multicolumn{1}{l|}{\begin{tabular}[c]{@{}c@{}} SSL methods \end{tabular}} & \multicolumn{1}{c}{bace} & bbbp & clintox & hiv & sider & tox21 & toxcast & Average\\ \midrule
Non-Pretrain & 72.80 $\pm$ 2.12 & 82.13 $\pm$ 1.69 & 74.98 $\pm$ 3.59 & 73.38 $\pm$ 0.92 & 55.65 $\pm$ 1.35 & 76.10 $\pm$ 0.58 & 63.34 $\pm$ 0.75  & 71.19\\ \midrule
ContextPred & 73.02 $\pm$ 2.59 & 80.94 $\pm$ 2.55 & 74.57 $\pm$ 3.05 & 73.85 $\pm$ 1.38 & 54.15 $\pm$ 1.54 & 74.85 $\pm$ 1.28 & 63.19 $\pm$ 0.94  & 70.65 (\ -0.54)\\
InfoGraph & 76.09 $\pm$ 1.63 & 80.38 $\pm$ 1.19 & \textbf{78.36 $\pm$ 4.04} & 72.59 $\pm$ 0.97 & \textbf{56.88 $\pm$ 1.80} & 76.12 $\pm$ 1.11 & 64.40 $\pm$ 0.84 & 72.11 (+0.93)\\
GPT-GNN & 75.56 $\pm$ 2.49 & 83.35 $\pm$ 1.70 & 74.84 $\pm$ 3.45 & 74.82 $\pm$ 0.99 & 55.59 $\pm$ 1.58 & 76.34 $\pm$ 0.68 & 64.76 $\pm$ 0.62 & 72.18 (+0.99)\\
GROVER & 75.22 $\pm$ 2.26 & 83.16 $\pm$ 1.44 & 76.8 $\pm$ 3.29 & 74.46 $\pm$ 1.06 & 56.63 $\pm$ 1.54 & 76.77 $\pm$ 0.81 & 64.43 $\pm$ 0.8 &72.5 (+1.31)\\
GraphCL & 76.47 $\pm$ 2.75 & 82.76 $\pm$ 1.00 &  77.74 $\pm$ 4.38 & 75.02 $\pm$ 1.03 & 56.12 $\pm$ 1.12 & 76.1 $\pm$ 1.34 & 63.25 $\pm$ 0.53 & 72.49 (+1.3) \\
\midrule
\textit{\method} & \textbf{77.22 $\pm$ 2.02} & \textbf{84.38 $\pm$ 1.07} & 77.02 $\pm$ 1.96 & \textbf{75.07 $\pm$ 1.07} & 56.67 $\pm$ 0.88 & \textbf{77.04 $\pm$ 0.77} & \textbf{65.23 $\pm$ 0.82} & \textbf{73.23 (+2.04)}\\
\bottomrule
\end{tabular}
}
\label{tab:finetune}
\end{table*}

\begin{table*}[h]
\small
\caption{Feature extraction performance (ROC-AUC) of \textit{\method} compared with other self-supervised learning (SSL) baselines on molecule property prediction benchmarks. Use pre-trained models to extract graph embeddings and train linear classifiers for ten times.}
\center
\resizebox{\textwidth}{!}{
\begin{tabular}{@{}l|ccccccc|c@{}}
\toprule
\multicolumn{1}{l|}{\begin{tabular}[l]{@{}c@{}}SSL methods\end{tabular}} & \multicolumn{1}{c}{bace} & bbbp & clintox & hiv & sider & tox21 & toxcast  & Average\\ \midrule
ContextPred & 53.09 $\pm$ 0.84 & 55.51 $\pm$ 0.08 & 40.73 $\pm$ 0.02 & 53.31 $\pm$ 0.15 & 52.28 $\pm$ 0.08 & 35.31 $\pm$ 0.25 & 47.06 $\pm$ 0.06 & 48.18 \\
InfoGraph & 66.06 $\pm$ 0.82 & 75.34 $\pm$ 0.51 & \textbf{75.71 $\pm$ 0.53} & 61.45 $\pm$ 0.74 & 54.70 $\pm$ 0.24 & 63.95 $\pm$ 0.24 & 52.69 $\pm$ 0.07 & 64.27 \\
GPT-GNN & 59.43 $\pm$ 0.66  & 71.58 $\pm$ 0.54 & 62.78 $\pm$ 0.58 & 64.08 $\pm$ 0.36 & 54.67 $\pm$ 0.16 & 68.20 $\pm$ 0.14 & 57.06 $\pm$ 0.13 & 62.53\\
GROVER & 65.67 $\pm$ 0.38 & 78.47 $\pm$ 0.36 & 53.19 $\pm$ 0.68 & 69.03 $\pm$ 0.23 & 54.94 $\pm$ 0.12 & 67.63 $\pm$ 0.13 & 57.28 $\pm$ 0.05 & 63.74\\ 
GraphCL & 60.93 $\pm$ 1.72 & 76.91 $\pm$ 0.85 & 73.79 $\pm$ 2.13 & 70.52 $\pm$ 1.22 & 55.01 $\pm$ 1.43 & \textbf{72.14 $\pm$ 0.78} & 58.51 $\pm$ 0.39 & 66.83 \\
\midrule
\textit{\method} & \textbf{70.83 $\pm$ 2.06} & \textbf{82.97 $\pm$ 1.53} & 73.49 $\pm$ 2.16 & \textbf{73.34 $\pm$ 1.27} & \textbf{57.32 $\pm$ 0.62} & 71.82 $\pm$ 0.82 & \textbf{59.46 $\pm$ 0.25} & \textbf{69.73}\\
\bottomrule
\end{tabular}
}
\label{tab:linear}
\end{table*}

We evaluate the effectiveness of \textit{\method} from two perspectives: 1) whether the self-supervised framework can learn better GNNs that generalize well on downstream graph classification tasks; 2) whether the learned motifs are reasonable and can genuinely benefit contrastive learning. 

We mainly focus on chemical property prediction tasks in this paper, where large-scale unlabelled molecules are available, and many downstream tasks are label-scarce. Specifically, we pre-train GNNs using \textit{\method} on the ogbg-molhiv dataset from Open Graph Benchmark (OGB) \citep{hu2020ogb}, which contains 40K molecules. We test our pre-trained model on smaller ogbg molecule property prediction benchmarks. For more details of the datasets, please see Appendix \ref{appendix:dataset}.

\subsection{Baselines and Model Configuration}
We consider six baselines, including non-pretrain (direct supervised learning) and five state-of-the-art GNN self-supervised learning (SSL) methods.

\textbf{InfoGraph} \citep{sun2019infograph} maximizes the mutual information between the representations of the whole graphs and the representations of its substructures.

\textbf{Context prediction} \citep{hu2020pretraining} predicts the surrounding structure of each node, so nodes appearing in similar structural contexts will be mapped to nearby representations.

\textbf{GPT-GNN} \citep{gpt_gnn} predicts masked edges and masked node attributes. The edge prediction makes node representations to be close when there are edges between them. The attribute prediction captures how node attributes are distributed over all graphs.

\textbf{GROVER} \citep{rong2020selfsupervised} first uses professional software, e.g. RDKit\citep{landrum2006rdkit}, to extract functional groups (motifs) from a dataset. Then, it pretrains by predicting motif labels.

\textbf{GraphCL} \cite{you2020graph} performs contrastive learning and construct views with four types of augmentations methods including node dropping, edge perturbation, attribute masking, and subgraph sampling. In our experiments, we adopt the default setting, i.e., randomly choose two out of four methods to construct views.


We use the state-of-the-art GNN model, Deeper Graph Convolutional Networks (DeeperGCNs) proposed in \citet{li2020deepergcn}, as the base GNN encoder for \textit{\method} and all baselines. We use the same hyperparameters for all experiments. Details about hyperparameters and model configurations are in Appendix \ref{appendix:hyper}.

\begin{table}[h]
\center
\small
\vskip -0.1in
\caption{Average feature extraction result with different ablations.}
\begin{tabular}{c|l|c}
\toprule
Ablations                          & Model Component            & Avg.  \\ \midrule
\multicolumn{2}{c|}{\method\ (Default Settings)}  & 69.73 \\ \midrule
\multirow{3}{*}{Subgraph Samplers} & Random Walk          & 64.09 \\ 
                                   & K-hop Sampler        & 64.74 \\
                                   & w/o ${\mathcal L}_{reg}$        & 67.91 \\ \midrule
\multirow{3}{*}{Contrastive Views} & Sub-Sub  (MICRO-G)            & 65.71 \\
                                   & Sub-Sub (GraphCL)    & 66.83 \\
                                   & Node-Sub (InfoGraph) & 64.27 \\ \midrule
   Subgraph Encoding & w/o contextualized emb & 61.68  \\ \midrule                             
\multirow{4}{*}{Number of Motifs}  & K = 5                & 67.96 \\ 
                                   & K = 10               & 68.89 \\
                                   & K = 20 (default)              & 69.73 \\
                                   & K = 50                 & 68.59 \\
                                   & K = 200                 & 68.87 \\
                                   \bottomrule
\end{tabular}
\vskip -0.1in
\label{tab:ablation}
\end{table}
\subsection{Evaluation Results under Different Protocols}\label{subsec:exp_results}
We evaluate the effectiveness of pre-trained GNNs using the following two evaluation protocols.

\textbf{Transfer Fine-tune Setting} To mimic the real-world setting with scarce data labels, we fine-tune the pre-trained GNN model on a small labeled data portion on downstream tasks. We adopt the same train-test and model selection procedure as in \citet{yanardag2015deep,  zhang2018end, xu2018powerful}, where we perform 10-fold cross-validation and report the epoch with the best cross-validation performance averaged over the 10 folds. The evaluation metric we used is the ROC-AUC score. 

\textbf{Feature Extraction Setting}: the setting is almost the same as fine-tuning, except that we fix the pre-trained GNN, use it as a feature extractor to get graph representations of all the datas, and train a linear classifier on top.

The evaluation results under transfer fine-tune setting and feature extraction setting are illustrated in Table \ref{tab:finetune} and \ref{tab:linear}. For both settings, \textit{\method} outperforms all baselines on average performance and achieves the highest results on most datasets. For the transfer fine-tune, we gain about 2.04$\%$ performance enhancement against the  non-pretrain baseline. For the feature extraction, \textit{\method} outperforms the strongest baseline GraphCL by 2.9$\%$. In addition, in the bbbp and sider datasets, the feature extraction performance of \textit{\method} without any fine-tuning outperforms the supervised baseline without pre-training. 


\subsection{Ablation Study} \label{subsec: ablation}
We conduct a series of ablation studies to analyze how different modules contribute the contrastive learning.
\paragraph{Motif-like Subgraph Sampling.} \label{ablation:sample}
Sampling subgraphs is a pervasive operation in graph learning. Simple approaches that often show up in literature include random walk and k-hop neighbours. The problem with these two sampling methods is that they only use local graph structural information but not feature information. Thus, they cannot accurately generate semantically-meaningful subgraphs when graph features are critical for representing the subgraph and whole-graph properties. 

\method\ leverages learned motifs to produce motif-like subgraphs. To evaluate its effectiveness, we run our framework by replacing motif-guided sampling with random walk and k-hop sampler, with all other settings stay the same. We show the result in the first block of Table \ref{tab:ablation}. As we see, the performance of the same framework with random walk and k-hop drops 5.64$\%$ and 4.99$\%$ respectively. This shows the importance of generating motif-like subgraphs. 


We also conduct ablation by removing the spectral regularizer ${\mathcal L}_{reg}$, and the performance drops 1.82$\%$. After removing this regularization, the sampled subgraphs are prone to contain disconnected nodes, which hinders the generalization performance to real-world graphs.

\textbf{Graph-to-Subgraph Contrastive} In this paper, we mainly utilize graph-to-subgraph (graph-sub) contrastive. Similar to image crop pair in computer vision, subgraph-to-subgraph (sub-sub) contrastive is also a promising alternative, which has already been studied in GraphCL \cite{you2020graph} with random walk. Thus, we try another ablation study by conducting sub-sub contrastive learning utilizing our motif-like subgraphs with all other settings unchanged. This setting's result is much worse than graph-sub learning, and worse than the GraphCL baseline.

We hypothesize that the performance gap of sub-sub with motif-like subgraphs is due to the false-negative of the current graph partition procedure. By grouping nodes close to a limited number of motifs, there is a high probability that two different graphs have a similar subgraph structure, which would form a false negative pair. To overcome this limitation, it is necessary to incorporate more randomness and take the composition of motifs as subgraph, which we leave these possibilities as future directions.

One might also ask why the graph-sub setting is less sensitive to  false-negatives. For our model design, one key choice is to first get all the contextualized node embeddings that encode the whole graph characteristics. In this way, even if two subgraphs from different graphs share a similar structure, their subgraph embeddings can still encode different context information. To test this hypothesis, we remove contextualized embedding by re-encoding all the sampled subgraphs. As shown in the third block, the result in this setting is extremely poor, worse than all existing pre-training frameworks. This explains the importance of contextualized node embeddings for subgraph-level contrastive learning and explains why our proposed graph-sub setting works.



\textbf{Number of Motifs} The number of motif slots, $K$, is an essential hyperparameter in our motif learning framework. We thus conduct an ablation study with three different $K$ values, i.e., 5, 10, 20, 50, and 200. As illustrated in the last block of Table \ref{tab:ablation}, with different $K$ values, \textit{\method} shows consistent performance enhancement, while an intermediate value 20 gives the best result on average. 

\begin{figure}[ht]
\vskip 0.2in
\begin{center}
\centerline{\includegraphics[width=0.95\columnwidth]{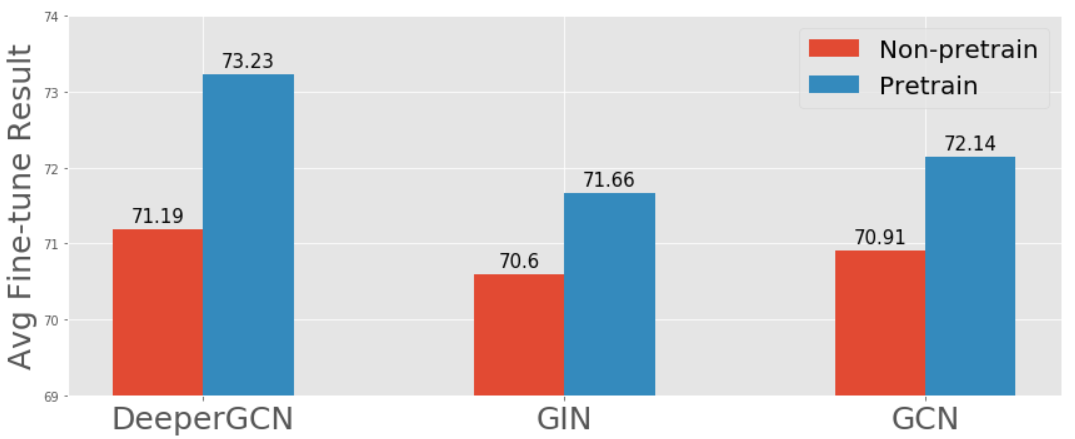}}
\caption{Average fine-tune result over three GNN architectures.}
\label{figure:GNNs}
\end{center}
\vskip -0.2in
\end{figure}

\textbf{GNN Architectures} The \textit{\method} framework is agnostic to the GNN architecture. We show this by trying three different standard GNN architectures, i.e. GCN~\cite{kipf2016semi}, GIN~\cite{xu2018powerful}, and DeeperGCN~\cite{li2020deepergcn}. As we showed in Figure \ref{figure:GNNs}, \textit{\method} provides a consistent performance enhancement for all different GNNs, and a more expressive GNN like DeeperGCN can provide a larger performance improvement.

\label{subsec:ablation}

\subsection{Visualization of learned Motifs}
\begin{figure}[ht]
\vskip 0.2in
\begin{center}
\centerline{\includegraphics[width=0.9\columnwidth]{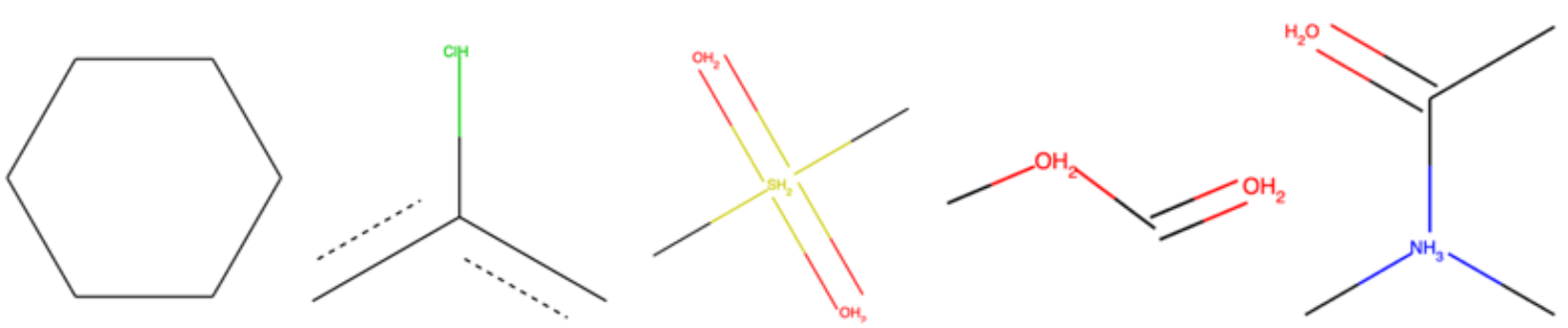}}
\caption{Five frequently occurred motifs learned by \textit{\method}, represented by their closest subgraphs.}
\label{figure:motif}
\end{center}
\vskip -0.2in
\end{figure}

We further show learned motifs by collecting the closest subgraphs to them. As illustrated in Figure \ref{figure:motif}, \textit{\method} automatically learns motifs that are similar to meaningful functional groups in the molecule domain, such as Benzene rings and acetate. This shows that \textit{\method} can learn reasonable and meaningful motifs. A complete list of the learned motifs is shown in Appendix \ref{appendix:topk}.

\section{Conclusion}\label{sec:conclusion}
We propose \textit{\method} to pre-train a GNN via subgraph-level contrastive learning. To tackle the challenge of informative subgraph sampling, we learn motifs during pre-training. With \textit{\method}, we learn meaningful motifs that align with molecular functional groups. Fine-tuning the pre-trained GNN on seven chemical property prediction benchmarks yields 2.04\% average improvement over non-pretrained GNNs and outperforms pre-training baselines.

\bibliographystyle{ACM-Reference-Format}
\bibliography{reference.bib}

\clearpage
\appendix
\section{Details of Motif Learning Framework}
\subsection{Necessity of Balance Regularization}
In Section \ref{sec:methodology}, we solve the node-to-motif assignment $\mQ^*$ using Eq (\ref{equation:sinkhorn_obj}). We now follow the probabilistic model framework to show this $\mQ^*$ is a reasonable choice in details.

\begin{figure}[b!]
\begin{center}
\includegraphics[width=0.95\columnwidth]{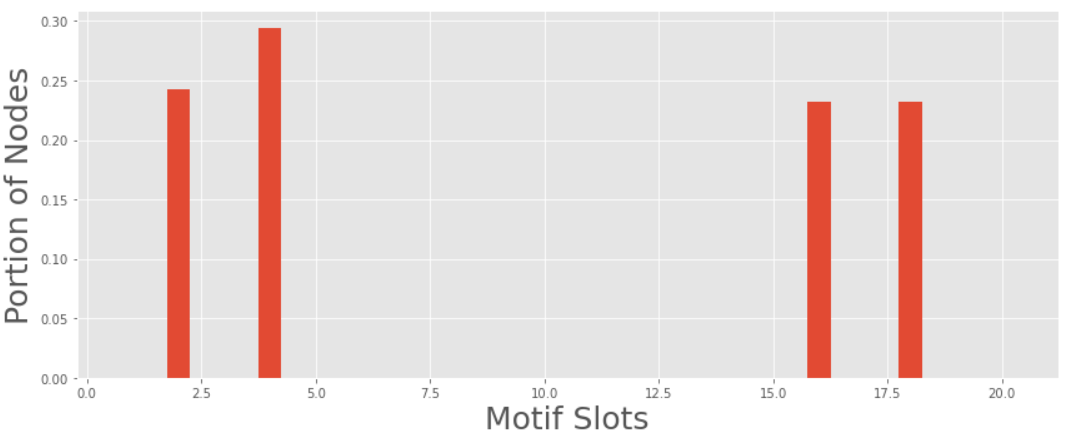} 
\end{center}
\vskip -0.2in
\caption{\textbf{Without} balanced regularization ($\mQ$): Distribution of number of assigned nodes of all the learned motifs.}
\label{figure:motif_unbalanced}
\end{figure}

For a standard probabilistic latent variable model with the EM algorithm, e.g. PLSA topic model, we compute the posterior probability $P(c_l = k \vert \vh_l)$ in the E-step. Then we use $\bm Q = P(c | u)$ as the reference distribution for deriving the lower bound of $\hat P(\gG \vert \bm{Par}, \mM, \theta)$ in the M-step. However, this approach has degenerate optimal solutions as we discussed in Section \ref{sec:methodology}, where all the nodes in a batch of graphs maybe assigned to only a few motifs. To show such situation, we use this vanilla assignment estimation $\bm Q$ without the balanced regularization to train the model. We quantify the motif balance by calculating the distribution of the number of nodes assigned to each motifs. As we can see from Figure~\ref{figure:motif_unbalanced}, the distribution is very unbalanced, where only few motifs have nodes assigned to it, and most motifs don't have any assignments. Such degeneration solution prevents the update of motif embeddings for those empty slots and significantly influences the model performance.





Therefore, instead of directly using $\mQ$ as the output of the E-step, we consider adding a regularization to balance assignments, and get the a regularized version $\bm\hat\mQ$ as the output. To calculate $\bm\hat\mQ$, we need to fulfill two requirements. The first is that $\mQ$ and the optimal $\bm\hat\mQ$ should tend to be similar; the second is that node assignments should be well spread over all motif slots to avoid the degenerate solutions. To achieve these two requirements, we consider the optimization problem in Eq (\ref{equation:sinkhorn_obj}).
\begin{gather} \label{equation:sinkhorn_obj2}
    \max_{\bm\hat \mQ \in \cal{Q}} Tr(\bm\hat\mQ \mQ^T) + \frac{1}{\lambda} H(\bm\hat\mQ) \text{,\  where}\\
    {\cal Q} = \{\bm\hat\mQ \in \mathbb{R}_+^{N_B, K} | \bm\hat\mQ \bm{1_{K}} = \frac{\bm{1_N}}{N}, \bm\hat \mQ^T \bm{1_{N}} = \frac{\bm{1_K}}{K} \} \nonumber
\end{gather}

Eq (\ref{equation:sinkhorn_obj2}) is a copy of the Eq (\ref{equation:sinkhorn_obj}). The first trace term of this objective is meant to make the estimate $\bm\hat\mQ$ close to $\mQ$. The second entropy term $H(\bm\hat\mQ)$ and the constrained search space $\cal{Q}$ are for regularization purpose. They will make $\bm\hat\mQ$ spread over different motifs, which can be seen as a inductive bias we add to the model. 

We pick this particular form for two reasons, one is that as shown in \cite{asano2020self}, solving this objective gives a good approximation of $\mQ$. More importantly, this objective is equivalent to an optimal transportation problem, which has a closed form solution and can be solved efficiently using a Sinkhorn-Knopp algorithm. We refer the readers to \cite{cuturi2013sinkhorn} for a detailed proof.

\begin{figure}[b!]
\begin{center}
\includegraphics[width=0.98\columnwidth]{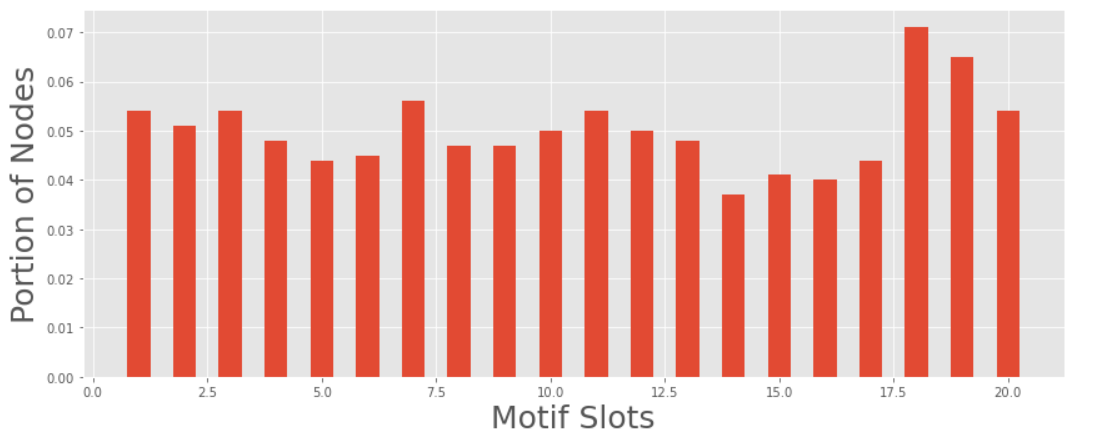} 
\end{center}
\vskip -0.2in
\caption{\textbf{With} balanced regularization ($\bm\hat\mQ$): Distribution of number of assigned nodes of all the learned motifs}
\label{figure:motif_balanced}
\end{figure}

We also show the node distribution using the balance regularized version $\bm\hat\mQ$ for assignment estimation. As shown in Figure~\ref{figure:motif_balanced}, the number of nodes in all motifs are very balanced, but still quite different from purly uniform as the regularization is just a soft constraint. This regularization well preserve the information of $\mQ$ while addressing the degenerate solution problem.

\subsection{Theoretical Justification of The EM Algorithm} \label{appendix:theory}

After solving Eq (\ref{equation:sinkhorn_obj2}) for the optimal $\bm\hat\mQ$, we get the discrete version $\mQ^*$ by only keeping the maximum column as in Eq (\ref{equation:discretization}). Then we use the discrete $\mQ^*$ to solve for the optimal partition $\bm{Par}^*$ in the M-step as in Eq (\ref{equation:subgraph}).

\begin{align} \label{equation:discretization}
\vq_{l,k}^* =  \begin{cases} 
1 &\text{k = $\argmax_k \hat \vq_{l,k}$}\\
0 &\text{otherwise}
\end{cases}  
\end{align}

The reason why we want to consider the discrete version is because solving $\bm{Par}^*$ is a discrete optimization problem. Doing partition requires us to have a hard assignment label of each node instead of a soft label. Although discretization makes the estimate less optimal, it greatly simplifies the problem and speeds up the search over all $\bm{Par}$ to be linear time. We would also like to point out that a similar approach of utilizing discrete assignments has been shown in the VQ-VAE paper Eq (1). \cite{van2017neural}. 

We now show the objective we optimized for the $\bm{Par}^*$. Here we look at the likelihood of a single subgraph, $\hat P(\vs_j \vert \bm{Par}, \mM, \theta)$. We assume uniform prior for all node assignments $c_l$ and subgraph assignments $z_j$.
\begin{align}\label{equation:likelihood}
&\hat P(\vs_j \vert \bm{Par}, \mM, \theta)\\
&= \sum_{k=1}^{K} \prod_{\substack{{l=1} \\ \vh_l \in \vs_j }}^{N} P(\vh_l \vert c_l = k, \mM, \theta) P(z_j = k)\\
&= \sum_{k=1}^{K} \prod_{\substack{{l=1} \\ \vh_l \in \vs_j }}^{N} P(\vh_l \vert c_l = k, \mM, \theta) \frac{1}{K}\\
&= \sum_{k=1}^{K} \prod_{\substack{{l=1} \\ \vh_l \in \vs_j }}^{N} P(\vh_l, c_l = k  \vert \mM, \theta)\\
&\propto \sum_{k=1}^{K} \prod_{\substack{{l=1} \\ \vh_l \in \vs_j }}^{N} P(c_l = k  \vert \vh_l, \mM, \theta) \\
&\approx \sum_{k=1}^{K} \prod_{\substack{{l=1} \\ \vh_l \in \vs_j }}^{N} \vq_{l, k}^*
\end{align}

\begin{algorithm*}[!h] 
\lstset{style=style_full}
\begin{lstlisting}[language=Python]
def __init__(self, args, encoder):
    super(MICRO_Graph, self).__init__()
    self.enc   = GNN(args) # Any GNN Model could be used
    self.M     = nn.Parameter(args.n_motifs, args.n_hid) #Motif Embedding Table
    self.W_h   = nn.Parameter(args.n_hid, args.n_hid)
    self.W_s   = nn.Parameter(args.n_hid, args.n_hid)
    self.W_e   = nn.Parameter(args.n_hid, args.n_hid)
    self.tau   = args.tau
    self.alpha = args.alpha
    
def forward(self, data):
    h, e = self.enc(data) # node and graph embeddings
    # Get Motif-like subgraphs via Partition
    # Detach self.M to stop grad flow for Motif here 
    Q = cos_sim(self.W_h*h, self.M.detach())
    Q = torch.softmax(Q / self.tau, dim=-1) # N_B x K
    with torch.no_grad():
        # Don't store gradient for discrete ops
        Q_hat = sinkhorn(Q)
        s, P_hat, num_subs = pool_sub(h, Q_hat) 
        # num_subs: number of subgraphs sampled from each G_i. 
        
    # Calculate motif-to-subgraph score. Detach GNN to avoid degeneration solution.
    P = cos_sim(self.W_s * s.detach(), self.M)
    P = torch.softmax(P / self.tau, dim=-1) # J_B x K
    
    # Calculate the two loss via M-step. 
    loss_m = self.motif_loss(Q, Q_hat, P, P_hat, data.adj)
    loss_c = self.contra_loss(s, e, num_subs)
    return self.alpha * loss_m + (1 - self.alpha) * loss_c
    
def motif_loss(self, Q, Q_hat, Q, P, P_hat, adj):
    loss_mot_sub  = - (P_hat * P.log()).sum(dim=1).mean() # Eq (13)
    loss_node_mot = - (Q_hat * Q.log()).sum(dim=1).mean() # Eq (10)
    loss_reg = spectral_loss(Q, adj) # Eq (11)
    return loss_mot_sub + loss_node_mot + loss_reg
    
def contra_loss(self, s, e, num_subs):
    blocks = [torch.ones(1, n) for n in num_subs]
    Y_lab = torch.block_diag(*blocks)
    Y = pairwise_cosine_sim(self.W_e * e, s)
    Y = torch.softmax(Y / self.tau, dim=-1)
    loss_contra = - (Y_lab * Y.log()).sum(dim=1).mean() # Eq (15)
    return loss_contra
    
def sinkhorn(Q, num_iters=5, lamb=20):
    # Implementation adopted from https://github.com/facebookresearch/swav
    Q = Q.transpose(0,1)  # K x N_B 
    Q = torch.exp(Q*lamb)
    Q /= torch.sum(Q)
    u = torch.zeros(Q.shape[0])
    r = torch.ones(Q.shape[0])/Q.shape[0]
    c = torch.ones(Q.shape[1])/Q.shape[1]
    curr_sum = torch.sum(Q, dim=1)
    
    for it in range(num_iters):
        u = curr_sum
        Q *= (r/u).unsqueeze(1)
        Q *= (c/torch.sum(Q, dim=0)).unsqueeze(0)
        curr_sum = torch.sum(Q, dim=1)
        
    Q_hat = (Q/torch.sum(Q, dim=0, keepdim=True))
    Q_hat = F.one_hot(Q_hat.argmax(dim=0))
    return Q_hat.transpose(0,1) #  N_B x K 
\end{lstlisting}
\caption{Pytorch-Style Code of \method, Full Version}\label{algo:full}
\end{algorithm*}

\subsection{Implementation Details of Graph Partition and Subgraph Generation}\label{sec:partition}
We describe more details about how exactly do we use the discretized node-to-motif assignments to partition graph and generate subgraphs.

As we introduced before, we first partition a graph by assigning its nodes into different motif slots and then we group all the nodes in same motif slot as a subgraph. For this procedure, the input is one whole graph $\gG$ and its corresponding node-to-motif assignment $\vq^* \in \{0, 1\}^{N \times K}$. Our output will be $J$ motif-like subgraphs, which can be different for each graph and determined by the assignment $\vq^*$. 
We first find out how many nodes are assigned to each motif. As $\vq^*$ is discrete, we simply calculate the row-wise sum $\vr_k = \sum_{l} \vq_{l,k}^*$ as node counts. To prevent generating subgraphs with only very few nodes, we filter out those motifs that contains less nodes than a threshold of $\eta$, i.e. $\vr_k < \eta$. In this way, the total number of subgraphs we generate for this whole graph will be $J = \sum_k \mathds{1}\{\vr_k \geq \eta\}$.

After partition, we get subgraphs by grouping all the nodes assigned to each motif. Since all these nodes have high probability belonging to this motif, the generated subgraph is also motif-like. Naturally, there could be a concern about applying this procedure to large graphs. If an input graph contains the same motif multiple times e.g., two hydroxy groups, then all relevant nodes in these two motifs will be assigned to the same motif and grouped to the same subgraph. This part is different from the traditional discrete motif mining, but such design choice is reasonable for motif representation learning. As the model can learn that duplication of basic motif components will not influence the overall motif property and thus should get similar embeddings. This is viable as our mean pooling of node embeddings is duplication-invariant.
For example, one hydroxy group and two hydroxy groups both imply solubility they will have similar subgraph embeddings and be classified to the same motif.

The threshold $\eta$ is another parameter we use to control subgraph sampling. $\eta = 0$ means we generate a subgraph corresponding to the a motif as long as there is at least one node assigned to it. We can also increase the value $\eta$ and make it a large-size filter, so generated subgraphs will have at least $eta$ nodes in it. In practice, we set the default $\eta$ to be 4.

To further increase the randomness of subgraph generation, we choose to randomly remove some nodes in one subgraph or add adjacency nodes to that subgraph. We thus get both informative and also diverse subgraphs.

\begin{figure*}[!h]
\begin{center}
\includegraphics[width=0.85\textwidth]{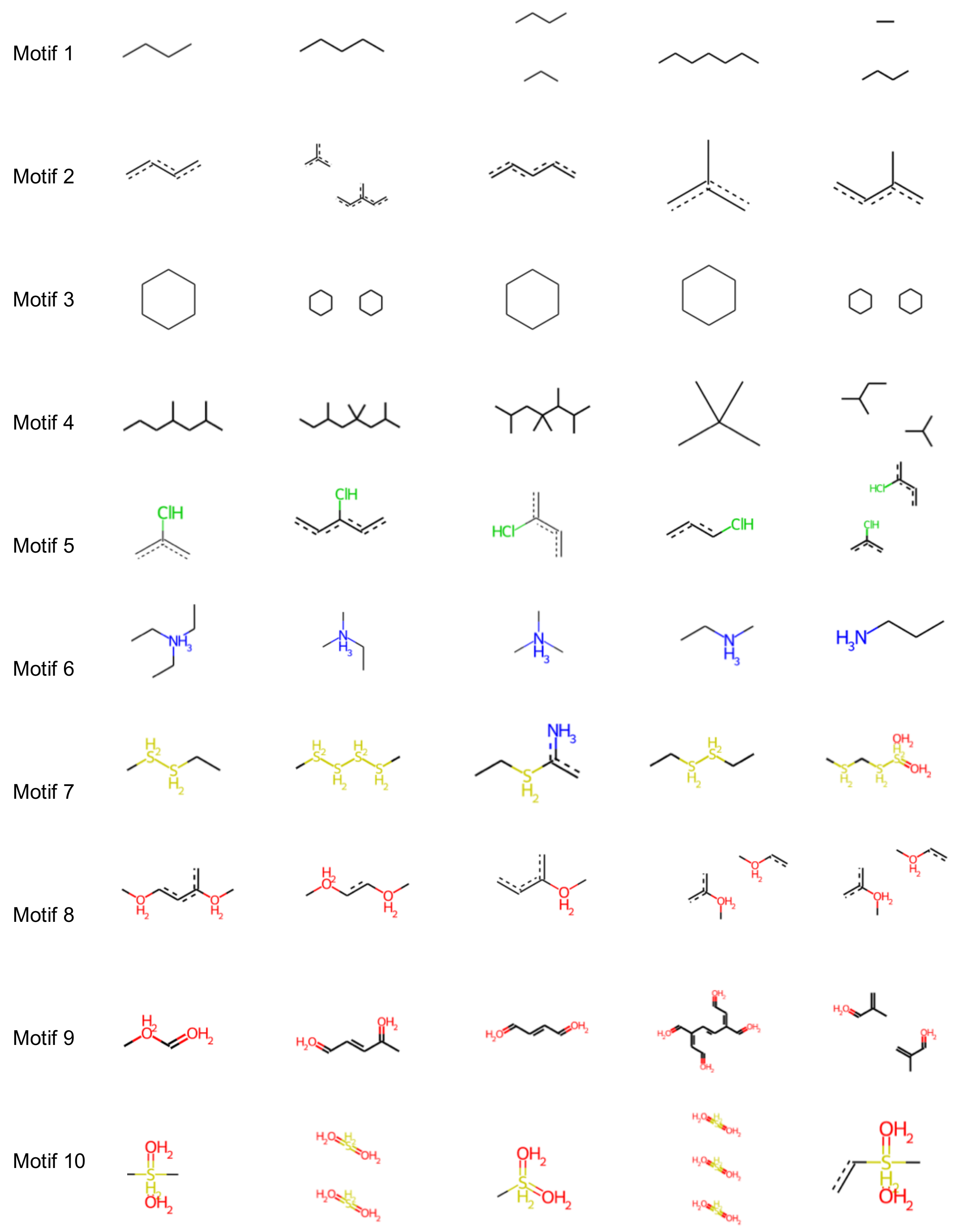} 
\end{center}
\caption{Motifs 1-10, represented by top k closest subgraphs to the learned motif representations. Each row corresponds to one motif, represented by some subgraphs that is closest to these motifs. Three columns indicate top 1 - 5 most similar subgraph respectively. }
\label{figure:topk1}
\end{figure*}

\begin{figure*}[!h]
\begin{center}
\includegraphics[width=0.87\textwidth]{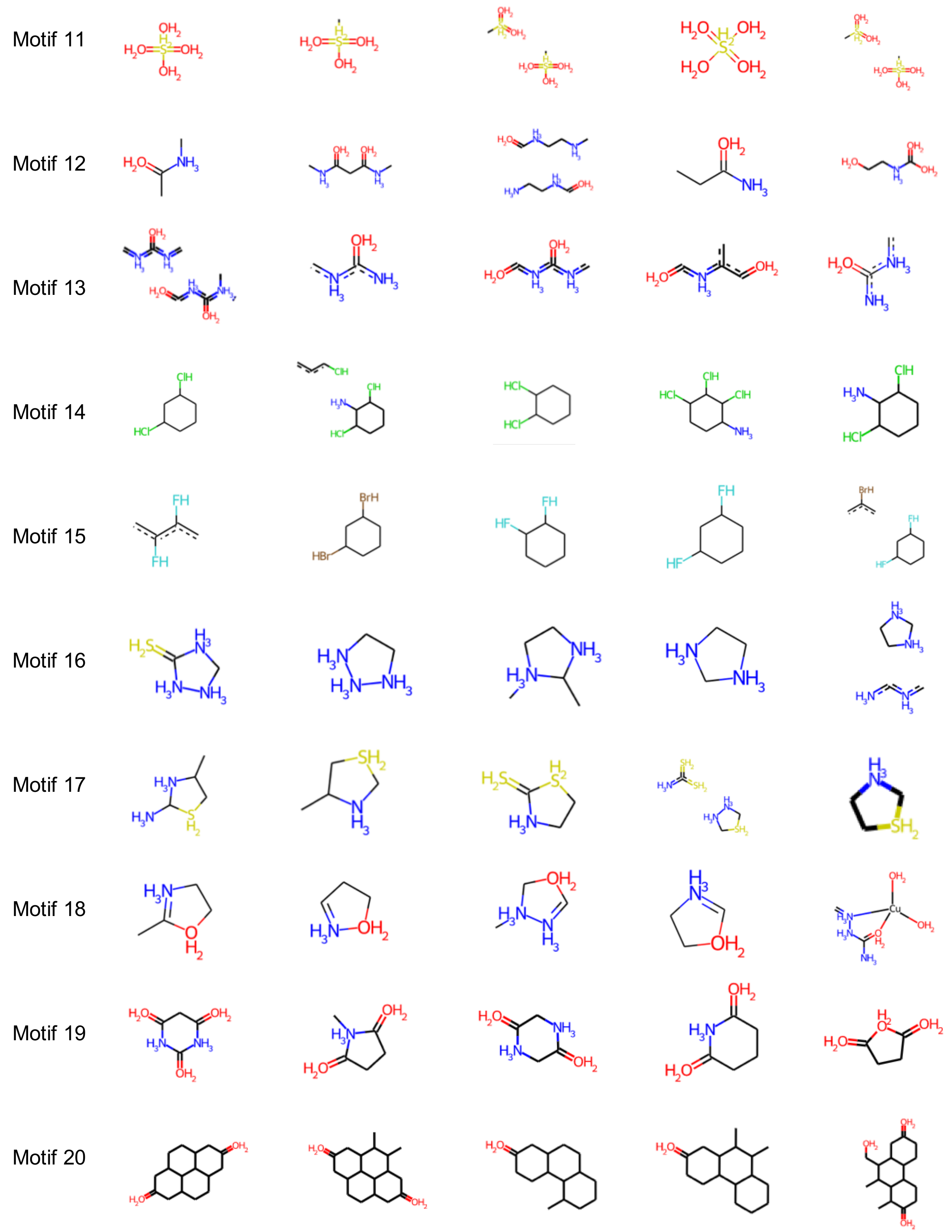} 
\end{center}
\caption{Motifs 11-20, represented by top k closest subgraphs to the learned motif representations. (Same setting as Figure \ref{figure:topk1})}
\label{figure:topk2}
\end{figure*}

\section{Pytorch-style Code of \method} \label{appendix:code}
We show the full pseudocode of our main algorithm in Algorithm \ref{algo:full}. Specifically, we first initialize the GNN encoder, Motif embedding table, and trainable projection parameter matrices $W$ in line 3-7. Note that Our algorithm is agnostic to GNN architecture, so any GNN model could be used. In our experiment, we considered GCN, GIN and DeeperGCN.

For a batch of input graphs $\{ \gG_1, \cdots, \gG_B\}$, i.e. \textit{data} in line 12, we first get the node embedding \textit{h} and whole graph embedding \textit{e}. Then we compute $\mQ$ according to Eq (\ref{equation:node_motif}) in line 17-20, and solve for the discreted and balanced $\mQ^*$ using the Sinkhorn-Knopp algorithm in line 16-19. Note that this step is for the E-step and involve non-differentiable steps, so we don't store gradient for it.

After that, we use $\mQ^*$ to partition the graph and group nodes into subgraphs according to section~\ref{sec:partition}, and then we pool the node embeddings to get subgraph embeddings \textit{s} in line 20. Plug these into the two loss functions gives us the final loss for optimizing both GNN encoder and Motif Embedding Table jointly in line 24-29.

\section{Motif Visualization and Interpretation}

\subsection{Top-k Closeset Subgraphs to Learned Motifs}
\label{appendix:topk}
To visualize the learned motif embeddings, we show the top-k similar subgraphs to each motif. The results of all the 20 learned motifs of the ogbg-molhiv dataset is shown in Figure \ref{figure:topk1} and \ref{figure:topk2}, which is the full version of Figure~\ref{figure:motif}. Since our learned motifs are vector embeddings, we interpret each motif by visualizing the subgraphs that are most similar to that the motif, e.g. for motif $\vm_k$, we visualize $g_j$ with the top $\vp_{j,k}$ values. 
We remove the subgraph duplicates that have exactly the same structure in top-k list to show more diverse results. We sort these motif by the average node size per subgraph, so the order reflects the complexity of the structure.


From the figure, We can make some interesting observations about the learned motif and subgraph embeddings: 

\begin{figure}[t!]
\begin{center}
\includegraphics[width=\columnwidth]{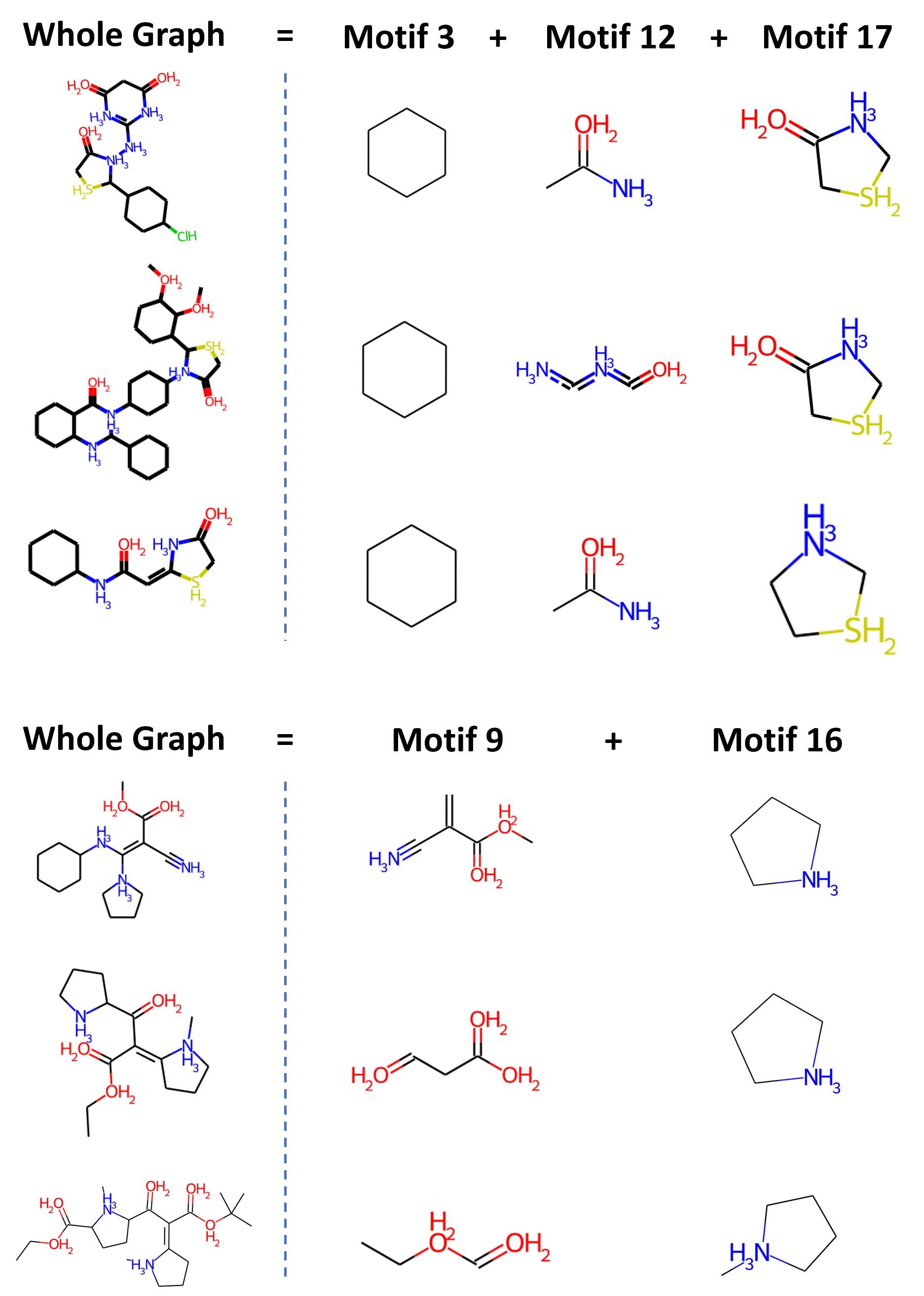} 
\end{center}
\caption{How the whole graphs could be represented by motif templates (combinations).}
\label{figure:comb} 
\end{figure}

\begin{figure*}[ht!]
\begin{center}
\includegraphics[width=\textwidth]{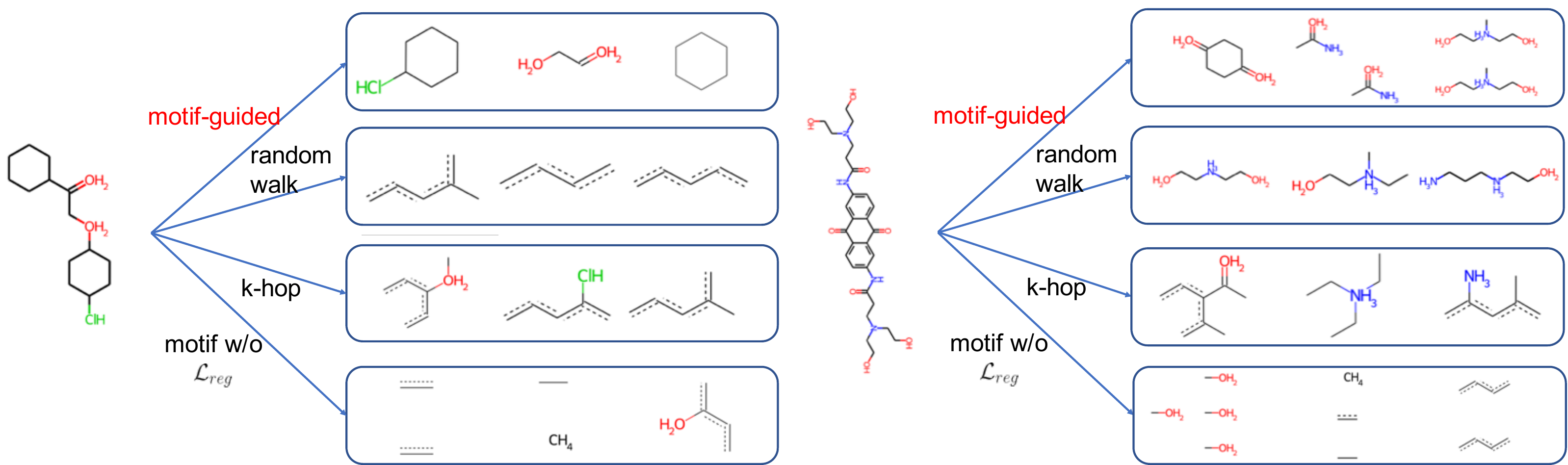} 
\end{center}
\caption{Comparison between different sampling strategies. Two examples are shown. In each example, the original graph is shown on the left. Samples produced by four different sampling strategies are shown on the right. The top row shows the samples by our motif-guided segmenter.}
\label{figure:four_sampling} 
\end{figure*}

\begin{enumerate}
    \item \textbf{The learned motif match real functional groups.} Even though our motifs are learned only from self supervision signals without any annotation or prior knowledge, we find that the learned molecule motifs match the real-world functional groups. For example, Motif 3 matches the benzene ring group\footnote{\url{https://en.wikipedia.org/wiki/Benzene}}, Motif 6 matches the Amine group\footnote{\url{https://en.wikipedia.org/wiki/Amine}}, Motif 9 matches carboxyl group\footnote{\url{https://en.wikipedia.org/wiki/Carboxylic_acid}}, Motif 10 matches sulfonyl group\footnote{\url{https://en.wikipedia.org/wiki/Sulfonyl}}, Motif 11 matches Sulfate\footnote{\url{https://en.wikipedia.org/wiki/Sulfate}}, Motif 14 and 15 matches 1,2- and 1,3-Dichlorobenzene group\footnote{\url{https://en.wikipedia.org/wiki/Dichlorobenzene}}. This shows that our motif learning can really capture the semantic property of molecule subgraphs, which is essential for the successful contrastive learning and model interpretation.
    
    \item \textbf{The learned motif can capture different level of property. } Similar to CNNs that learn to extract a hierarchy of visual patterns, from local patterns like edge and texture, to high-level patterns like concept, our learned motif also have different levels, even if we didn't add explicit hierarchical modelling. By sorting the motif via average node size, we can see that the simple motifs tend to model some frequently-occured building blocks such as carbon chain in Motif 1, carbon triangle in Motif 2, benzene ring in Motif 3, etc. While the complex motifs model sub-molecule or functional group, such as the Amino acid in Motif 12 and 13, Dichlorobenzene in Motif 14 and 15. Such hierarchy shows the diversity of our learned motif, and explain why it can fit a wide range of molecule property prediction tasks. This finding also motivates us to think about extending the current model to model the hierarchy of motif pattern, which we leave for future works.
    
    \item \textbf{The learned Motif can capture semantic property beyond structure.} As we discussed before, one of the limitation of traditional motif mining is that they only consider structure similarity, and thus even two structurally-different subgraphs share similar chemical property, they will not be regarded as the same motif. On the contrary, our soft motif embedding can capture high-level semantic beyond structure. Take the motif 15 as an example. We know that fluorine (F), chlorine (Cl), bromine (Br), iodine (I) are all halogens, which all belong to Group 17 in the periodic table. Therefore, all subgraphs with the pattern that two halogens linked to benzene ring adjacently share similar chemical properties and have very similar contexts. Our Motif Learning can successfully classify all these subgraphs into motif 15, even if their structure and node feature are different.

\end{enumerate}

\subsection{Motif Template (Combination) for whole graph}
We further show that the learned motif template could also be used to characterize the whole graph property. As the motif is the basic building blocks, each whole graph could have a corresponding motif vector representing which motif it contains. Therefore, we cluster over a batch of input graph to get some frequent motif combination. For example, in Figure~\ref{figure:comb}, we can see that there are two motif combination (3, 12, 17) and (9, 16) as templates. For each motif combination, we can find all the graphs that only contain these motifs. As is shown in the figure, for combination (3, 12, 17), although the three graphs are structurally different, they all contain some essential subgraphs that are associated to the corresponding motif. Therefore, their chemical properties are likely to be similar. Such analysis could help researchers to extract discrete features (the motif combination) to interpret model decision.

\section{Sampling Strategies and Sampled Subgraphs} \label{appendix:sampling}
One of the key component in our \method\ is the motif-guided sampling, and in the ablation study Section \ref{subsec: ablation} we compare it with two heuristic sampling strategies, i.e. random walk and k-hop. Here we describe the implementation details of the two heuristic samplers, and also visualize the sampled subgraphs under different strategies to give a more intuitive understanding of their limitations.

For random walk, we adapt the PyG's implementation\footnote{\url{https://github.com/rusty1s/pytorch_cluster/blob/master/torch_cluster/rw.py}}. We use a random walk length uniform in [10, 40]. Starting from a randomly selected seed node in the graph, we randomly select its neighborhood as next hop, until reaching the walk length threshold. Then the sample is the induced graph of all nodes have been visited.

For $K$-hop neighbors, we adapt theaw PyG's implementation\footnote{\url{https://pytorch-geometric.readthedocs.io/en/latest/_modules/torch_geometric/utils/subgraph.html}}. 
We pick hop number $k$ to be 1 or 2 with equal probability. Starting randomly selected seed node in the graph, we collect all the neighbors within $k$ hop as the sampled subgraph.

Two examples of subgraphs generated by all four strategies are shown in  Figure~\ref{figure:four_sampling}.

From the sampled subgraphs, we can see that random walk is more likely to generate chains (as it just randomly pick next-hop nodes, and thus have low probability to sample a complete benzene ring, which is the most basic component of molecule subgraph), while k-hop sampling is more likely to generate half part of a Benzene ring (as it just add in all the local neighborhood without selecting the important ones, and thus most subgraphs look similar). Neither of these two heuristic approaches can successfully generate a complete and clean functional group (such as benzene rings), and the generated subgraphs are not very meaningful. On the contrary, our motif-guided sampling can successfully generate a complete benzene rings and other significant substructures. This intuitively explains why the contrastive learning with our motif-guided sampler works much better than the others. After removing the ${\mathcal L}_{reg}$ term in the joint loss, the samples can still capture important nodes in the whole graph, but it will overlook some structural information and are prone to get some disconnected parts in a subgraph, which is not very realistic.

To sum up, our learned motif can indeed help generate more meaningful and diverse subgraphs, and thus help the contrastive learning to pre-train better GNN.

\begin{table}[!h]
\center
\resizebox{\columnwidth}{!}{
\begin{tabular}{@{}l|lllllll@{}}
\toprule
\multicolumn{1}{c|}{\begin{tabular}[c]{@{}c@{}}Dataset\end{tabular}} & \multicolumn{1}{c}{bace} & bbbp & clintox & hiv & sider & tox21 & toxcast \\ \midrule
\# graphs & 1513 & 2039 & 1477 & 41127 & 1427 & 7831 & 8576 \\
\# nodes & 51577 & 49068 & 38637 & 1049163 & 48006 & 145459 & 161088 \\
\# edges & 111536 & 105842 & 82372 & 2259376 & 100912 & 302190 & 161088 \\
\# tasks & 1 & 1 & 2 & 1 & 27 & 12 & 617 \\ \bottomrule
\end{tabular}
}
\caption{Statistics on number of graphs, nodes, edges, and tasks in each OGB molecule dataset.}
\label{tab:datastats}
\end{table}

\begin{table*}[h]
\caption{Summary of Experiment Settings and Hyper-parameters}
\center
\resizebox{\textwidth}{!}{
\begin{tabular}{@{}l|c|c@{}}
\toprule
\textbf{Name} & \textbf{Description} & \textbf{Value} \\ \midrule
\multicolumn{3}{c}{Optimization-related Hyperparameters} \\ \midrule
learning rate & learning rate for both pretraining and downstream & 1e-3\\ 
pre-train epochs & number of epochs for the pre-training stage & 200\\ 
pre-train batch size & batch size for the pre-training stage & 512\\ 
pre-train dropout & dropout ratio for the pre-training stage & 0.2\\ 
downstream epochs & number of epochs for both fine-tune and linear evaluation & 100\\ 
downstream batch size & batch size for both fine-tune and linear evaluation & 32\\ 
downstream dropout& dropout ratio for both fine-tune and linear evaluation & 0.5 \\ \midrule
\multicolumn{3}{c}{GNN Model Hyperparameters} \\ \midrule
conv\_name & Name of GNN conv layer for each architecture & GCN, GIN, DeeperGCN \\
n\_layers & number of GNN hidden layers & 5 \\
n\_hid ($d$) & dimension of node/graph/motif embeddings & 300 \\
pool & GNN graph pooling method & mean \\
norm & GNN normalization method & batch norm \\\midrule
\multicolumn{3}{c}{\textit{\method} Specific} \\ \midrule
$K$ & number of motifs & 5,10,20,50,200 \\
$\lambda$ & parameter before the entropy term in the Sinkhorn-Knopp algorithm, Eq (\ref{equation:sinkhorn_obj}) & 20 \\
proj\_dim & dimension of the projected embeddings, e.g. first dimension of $W_h$, $W_s$ and $W_e$ & 300 \\
$\tau$ & temperature parameter for softmax calculation & 0.05 \\ 
$\alpha$ & parameter balancing motif loss and contrastive loss, Eq (\ref{equation:joint_loss}) & 0.5 \\
$\lambda_n$ & coefficient of ${\mathcal L}_{node\text{-}mot}$, Eq  \ref{equation:motif_total_loss} & 1 \\
$\lambda_s$ & coefficient of ${\mathcal L}_{mot\text{-}sub}$, Eq  \ref{equation:motif_total_loss} & 1 \\
$\lambda_r$ & coefficient of ${\mathcal L}_{reg}$, Eq  \ref{equation:motif_total_loss} & 5 \\
$\eta$ & subgraph sampling filtering threshold & 4 \\
\bottomrule
\end{tabular}
}
\label{tab:hyperparameters}
\end{table*}

\section{Chemical Property Prediction Benchmarks}
\label{appendix:dataset}
In our experiments, we evaluated model performance on seven Open Graph Benchmark (OGB) molecule property prediction datasets. We provide a synopsis of each downstream task dataset taken verbatim from \cite{hu2020pretraining} below:

\begin{compactitem}
    \item \textbf{bace:} Qualitative binding results for a set of inhibitors of human $\beta$-secretase 1.
    \item \textbf{bbbp:} Blood-brain barrier penetration (membrane permeability).
    \item \textbf{clintox:} Qualitative data classifying drugs approved by the FDA and those that have failed clinical trials for toxicity reasons.
    \item \textbf{hiv:} Experimentally measured abilities to inhibit HIV replication.
    \item \textbf{sider:} Database of marketed drugs and adverse drug reactions (ADR), grouped into 27 system organ classes.
    \item \textbf{tox21:} Toxicity data on 12 biological targets, including nuclear receptors and stress response pathways.
    \item \textbf{toxcast:} Toxicology measurements based on over 600 in vitro high-throughput screenings.
\end{compactitem}

Table \ref{tab:datastats} summarizes important statistics of the OGB molecule datasets related to the number of graphs, the size of graphs, and number of properties that require prediction for each molecule. For these datasets, there are 9-dimensional node features including atomic number, chirality, and etc. There are also 3-dimensional edge features including bond type, bond stereochemistry, and an additional bond feature indicating whether the bond is conjugated. For further information on the OGB datasets, please refer to \cite{hu2020pretraining} and \cite{hu2020ogb}.

\section{Summary of Experiment Settings and Hyper-parameters} \label{appendix:hyper}

We release our code and data \footnote{\url{https://drive.google.com/file/d/1b751rpnV-SDmUJvKZZI-AvpfEa9eHxo9/}} for re-productivity. We also summarize the experiment settings and relevant hyper-parameters used for our experiments in Table \ref{tab:hyperparameters}. As we mentioned before, we use the same GNN architecture for \textit{\method} and all the baseline models. The optimization parameters and the GNN parameters showing in Table \ref{tab:hyperparameters} stay the same across all experiments.

We pre-train our model using Adam optimizer for 100 epochs, with batch size (number of graphs per batch) 512. For fine-tuning, we train the model for 100 epochs with batch size 32. We select model with highest validation result and report its test result. Corresponding experiment results are shown in Section \ref{subsec:exp_results}.

We didn't do too much hyper-parameters tuning, for optimization and GNN architecture, we use the recommended hyper-parameters in \cite{hu2020pretraining}. We tuned the important hyper-parameter for \textit{\method}, e.g. $K$, and discussed its influence in Section \ref{subsec: ablation}. For the parameters balancing different terms in the joint loss, e.g. $\lambda_n, \lambda_s, \lambda_r$, we also simply take their values to be equal to one, and we take $alpha = 0.5$. The final loss will be a direct summation of all the terms.

\begin{figure*}[!h]
\begin{center}
\includegraphics[width=\textwidth]{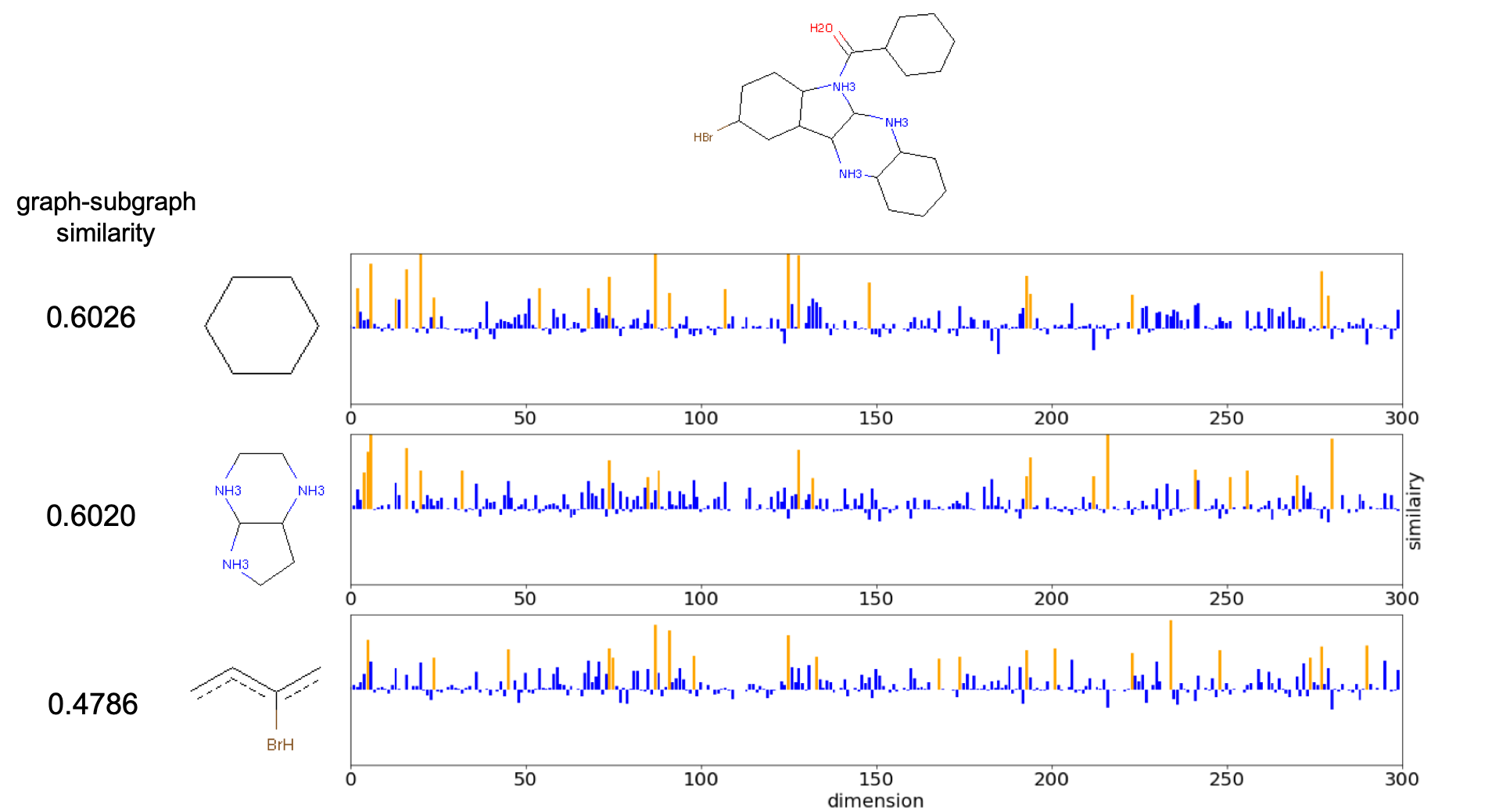} 
\end{center}
\caption{Similarity between the whole graph $\gG_1$ and three subgraphs $g_1$, $g_2$, and $g_3$, zoom in to each dimension. For each row, x-axis is the dimension slot 1 to 300, and y-axis is the similarity scores between corresponding dimensions of the whole graph representation and each subgraph representation. We indicate the top 20 scores in orange. We can see that these three subgraphs have very different similarity score distributions, though summing over all 300 dimensions give alike high scores.}
\label{figure:whole_sub_intra_sim}
\end{figure*}

\section{Case Study of Graph-to-Subgraph Contrastive and The Learned Embeddings}
\begin{table*}[ht!]
\center
\caption{Ablation study results. Full table.}
\resizebox{\textwidth}{!}{
\begin{tabular}{@{}l|ccccccc|c@{}}
\toprule
\multicolumn{1}{l|}{\begin{tabular}[l]{@{}l@{}}Sampler\end{tabular}} & \multicolumn{1}{c}{bace} & bbbp & clintox & hiv & sider & tox21 & toxcast  & Average \\ \midrule

RW & 66.09	$\pm$ 02.61& 76.16	$\pm$ 01.02& 61.21	$\pm$ 05.03& 69.69	$\pm$ 00.82& 53.74	$\pm$ 00.82& 65.03	$\pm$ 00.75& 56.72 $\pm$ 00.48& 64.09 \\ 
K-hop & 68.93	$\pm$ 01.24 & 74.36	$\pm$ 01.06 & 62.15	$\pm$ 03.96 & 70.06	$\pm$ 00.95 & 54.04	$\pm$ 00.67 & 67.21	$\pm$ 00.71 & 56.4 $\pm$ 00.5 & 64.74\\ 
w/o ${\mathcal L}_{reg}$ & 68.63 $\pm$ 2.18 & 79.54 $\pm$ 0.84 & 67.95 $\pm$ 2.24 & 71.7 $\pm$ 1.38 & 57.58 $\pm$ 1.13 & 71.57 $\pm$ 0.49 & 58.37 $\pm$ 0.34  & 67.91 \\ \midrule
Sub-Sub(MICRO-G) & 70.75 $\pm$ 2.91 & 80.64 $\pm$ 0.97 & 62.78 $\pm$ 5.27 & 70.83 $\pm$ 0.82 & 56.78 $\pm$ 0.66 & 63.95 $\pm$ 0.71 & 54.42 $\pm$ 0.43  & 65.71 \\ \midrule
w/o contextualized emb  & 60.11 $\pm$ 2.15 & 74.48 $\pm$ 1.11 & 51.39 $\pm$ 3.83 & 66.53 $\pm$ 1.06 & 54.63 $\pm$ 0.8 & 67.36 $\pm$ 0.62
 & 57.26 $\pm$ 0.53 & 61.68  \\ \midrule
K = 5 & 73.54  $\pm$ 1.85& 81.94$\pm$ 1.22& 65.08$\pm$ 2.35& 72.80$\pm$ 0.99& 57.14$\pm$ 0.62& 71.33$\pm$ 0.75& 59.52  $\pm$ 0.58 
 & 67.96 \\
K = 10 & 72.66	$\pm$ 2.92 & 80.07 $\pm$ 1.32	&68.12 $\pm$ 1.85&	73.25 $\pm$ 1.2 &	57.12 	$\pm$ 00.49 &71.69 $\pm$ 0.8&	59.37 $\pm$ 0.44 & 68.89 \\
K = 20 (default) & 70.83 $\pm$ 2.06 & 82.97 $\pm$ 1.53 & 73.49 $\pm$ 2.16 & 73.34 $\pm$ 1.27 & 57.32 $\pm$ 0.62 & 71.82 $\pm$ 0.82 & 59.46 $\pm$ 0.25 & 69.73 \\
K = 50 & 74.45 $\pm$ 2.24 &	81.46	$\pm$ 1.04 & 62.4	$\pm$ 2.08	& 73.35	$\pm$ 1.21 & 57.21	 $\pm$ 0.78 & 72.24 $\pm$ .8 &	59.11 $\pm$ 0.46 & 68.59 \\
K = 200 & 71.61 $\pm$ 1.08 & 84.25 $\pm$ 1.08 &	67.08 $\pm$ 3.2 &	73.53 $\pm$ 0.75 &	56.92 $\pm$ 1.28 &	72.46	$\pm$ 0.58 & 58.98 $\pm$ 0.5  & 68.87 \\ 
\bottomrule
\end{tabular} \label{tab:ablation_full}
}
\end{table*}

Another core component in \textit{\method} is the graph-to-subgraph contrastive learning based on motif-like subgraphs. Though we have previously showed that such design can empirically and intuitively help pre-traing better GNN, there's still some potential question about the combination of motif with contrastive learning. 

\subsection{Different subgraphs (motifs) from the same graph}
One key question is that each graph can be partitioned into subgraphs that belong to different motifs. Through graph-to-subgraph contrastive learning, we force these subgraph embeddings to be similar to the whole graph embedding. However, will these subgraph embeddings also be forced to be similar to each other? If so, it contradicts to our principle of learning distinctive motif semantics.

\begin{figure}[ht!]
\begin{center}
\includegraphics[width=0.7\columnwidth]{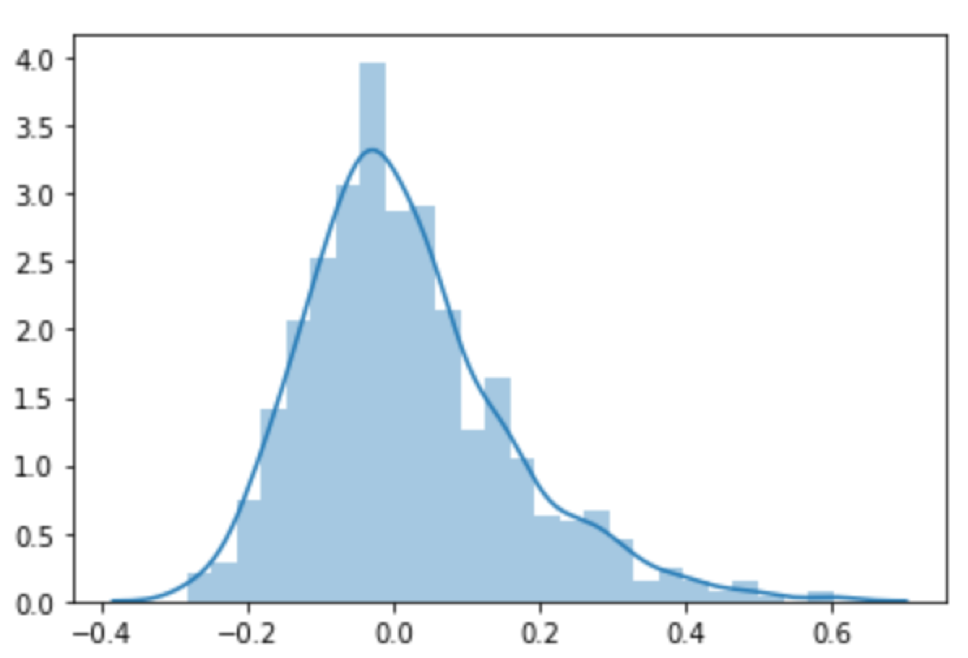} 
\end{center}
\caption{Distribution of similarity scores between the whole graph $G$ and all the subgraphs}
\label{figure:whole_sub_all_sim}
\end{figure}

We investigate this by a case study of a particular graph in Figure~\ref{figure:whole_sub_all_sim}, which is partitioned into three subgraphs that belong to different motif slots. We show the similarity score of these subgraph to the whole graph, and also the entry-wise similarity of each hidden dimension of the 300-dimension embedding. As we can see, all the three subgraphs can get relatively high similarity score, compared to the overall distribution of graph-to-subgraph similarity score shown in Figure \ref{figure:whole_sub_all_sim}. 

One very interesting findings is that the maximum entry of similar score for these three subgraphs are very different. Specifically, this indicates that the 300-dimension embedding actually encode multi-view semantic information. While doing cosine similarity of subgraph to whole graph, different dimension could be activated for different subgraphs. In other words, they are only similar to the projection of the whole graph representation on different basis. Therefore, even the three subgraphs are all similar to the whole graph, there embeddings are not collapse to be the same, which maintains the diversity and distinctiveness of the motifs.

\begin{figure}[ht!]
\begin{center}
\includegraphics[width=0.9\columnwidth]{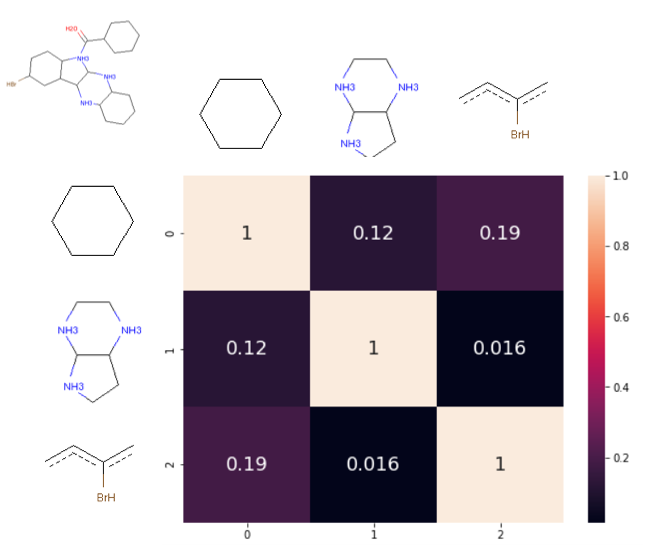} 
\end{center}
\caption{Pairwise similarity scores between subgraphs $g_1$, $g_2$, and $g_3$ from the same whole graph.}
\label{figure:sub_sim}
\end{figure}

To further justify our claim, we also show the pairwise cosine similarity scores between these three subgraphs in Figure \ref{figure:sub_sim}. We find that their mutual similarity is not very low, indicating that each subgraph embeddings could capture their own semantic. 

\begin{figure}[ht!]
\begin{center}
\includegraphics[width=0.7\columnwidth]{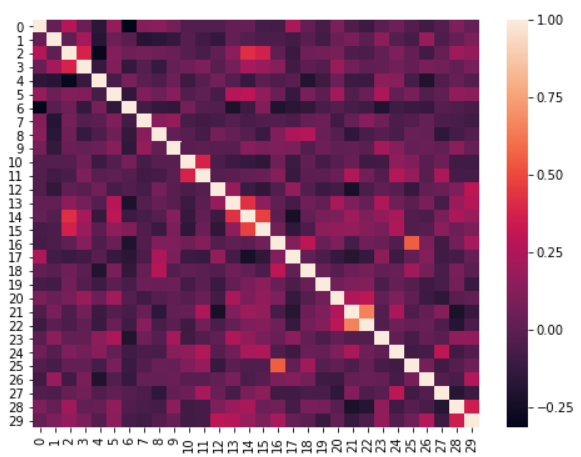} 
\end{center}
\caption{Pairwise similarity scores between the first 30 subgraphs sampled from $\{\gG_1,..., \gG_B\}$}
\label{figure:sub_sim2}
\end{figure}

To show that this claim is generalizable to the whole dataset, we randomly select a batch of graph $\{\gG_1,..., \gG_B\}$, and get all their subgraphs. We use Figure \ref{figure:sub_sim2} to show the pairwise similarity scores between these subgraphs in order. we find that even though these subgraphs are listed in order, (i.e. $g_1, ..., g_3$ are from $\gG_1$, $g_4, g_5$ are from $\gG_2$, $g_6, ... g_8$ are from $\gG_3$, and etc) similarity scores are roughly uniform. In other words, this heat matrix is not strictly block diagonal, indicating subgraphs from the same whole graph do not necessarily have high similarities among them. This again justifies that our claim is true among the whole dataset.

\begin{figure}[ht!]
\begin{center}
\includegraphics[width=0.9\columnwidth]{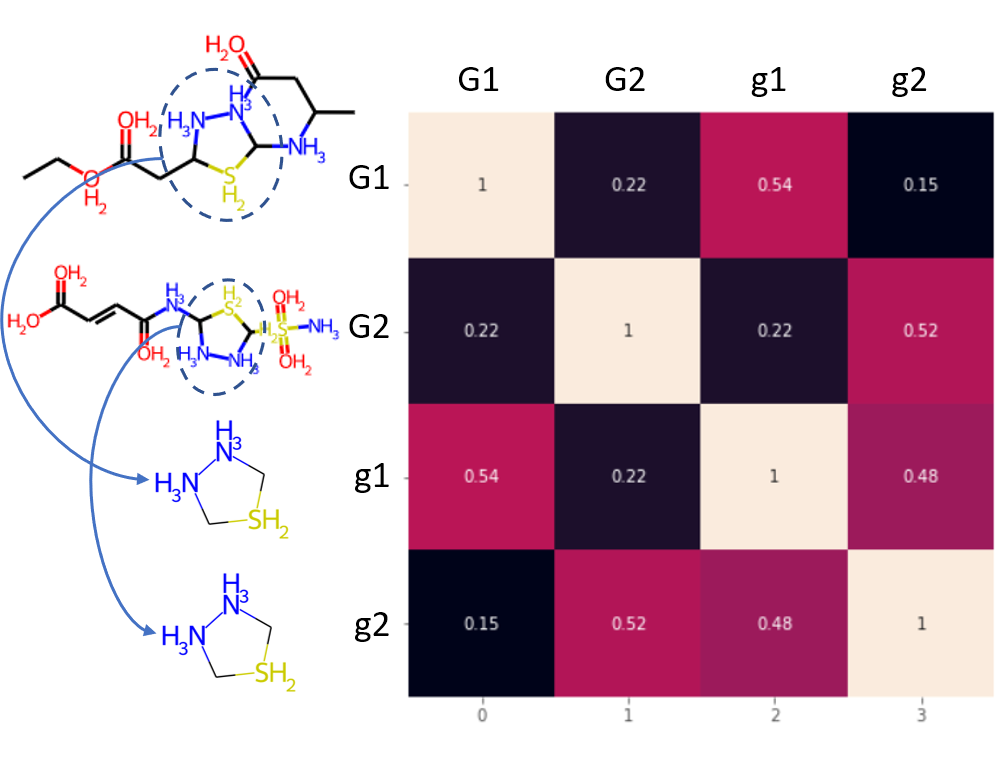} 
\end{center}
\caption{Pairwise similarity scores between Graph $G_1$, $G_2$ and their contained structurally-similar subgraph $g_1$ and $g_2$.}
\label{figure:same}
\end{figure}

Also, this study partially answers why graph-to-subgraph contrastive works better than subgraph-to-subgraph contrastive in our setting. Using subgraph-to-subgraph contrastive, two subgraph in the same graph that have different motif assignments will be forced to be similar, which contradicts to our assumption to motif embedding. If we want to extend the current framework to subgraph-to-subgraph contrastive, a better subgraph sampling that considers composition of motif and more randomness is required.

\subsection{Similar subgraphs (motifs) from different graphs}
A similar question people could ask is that: should the two subgraph of the similar motif but belongs to different graph be similar or different. In our setting, their subgraph embedding also encode the contextual neighborhood information, so even if their subgraph embedding is similar, we can still distinguish whether they belong to the correct graph. 

We show such a case in Figure~\ref{figure:same}, the two different whole graphs $G_1$ and $G_2$ contains exactly the same subgraph that belongs to Motif 17. 

The cosine similarity between $G_1$ and $g_1$ is 0.54, while similarity between $G_2$ and $g_2$ is 0.52, which fits our assumption for the positive contrastive. However, the similarity between $G_1$ and $g_2$ is 0.15, and similarity between $G_2$ and $g_1$ is 0.22, which are significantly lower than the positive pairs. If our subgraph embedding only encodes the structure information without considering contexts, the score should be the same. Therefore, due to the contextualized node representation, the subgraph embedding for $g_1$ and $g_2$ are different. This is also supported by the cosine similarity between $g_1$ and $g_2$, which is only 0.48 instead of 1. Therefore, our contrastive learning can handle such "same subgraph from different graph" case, and guide the model to capture global information.

\section{Full results of Ablation Studies over all datasets}
For the ablation studies in Sec \ref{subsec: ablation}, we only reported the average result due to space limit. We now list the full table of all the ablation experiments we have run in Table \ref{tab:ablation_full}. The results for each datasets are consistent and the analysis and conclusion are generalizable to them all.

\end{document}